%% file: neurips_2026_MedGround.tex
\documentclass[eandd]{article}

\usepackage[preprint]{neurips_2026}
 \usepackage{neurips_2026}

\usepackage[utf8]{inputenc} 
\usepackage[T1]{fontenc}    
\usepackage{hyperref}       
\usepackage{url}            
\usepackage{booktabs}       
\usepackage{amsfonts}       
\usepackage{nicefrac}       
\usepackage{microtype}      
\usepackage{xcolor}         

\usepackage{latexsym}
\usepackage{inconsolata}
\usepackage{graphicx}
\usepackage{amssymb}
\usepackage{pifont}
\usepackage{tabularx}
\usepackage{adjustbox}
\usepackage{multirow}
\usepackage{caption}
\usepackage{cleveref}
\usepackage{geometry}

\definecolor{posgreen}{HTML}{008000}
\definecolor{negred}{HTML}{FF0000}

\title{MedGround: Bridging the Evidence Gap in Medical Vision-Language Models with Verified Grounding Data}

\author{
 \textbf{Mengmeng Zhang\textsuperscript{1,2,3}},
 \textbf{Xiaoping Wu\textsuperscript{3}},
 \textbf{Hao Luo\textsuperscript{3,4$^{\dagger}$},
 \textbf{Fan Wang\textsuperscript{3}},
 \textbf{Yisheng Lv\textsuperscript{1,2$^{\dagger}$}}
 }
\\
 \textsuperscript{1}Institute of Automation, Chinese Academy of Sciences, \\
 \textsuperscript{2}School of Artificial Intelligence, University of Chinese Academy of Sciences, \\
 \textsuperscript{3}DAMO Academy, Alibaba Group, \\
 \textsuperscript{4}Hupan Lab, Zhejiang Province \\
 \small{
   \textbf{Correspondence:}
   \href{mailto:michuan.lh@alibaba-inc.com}{michuan.lh@alibaba-inc.com},
   \href{mailto:yisheng.lv@ia.ac.cn}{yisheng.lv@ia.ac.cn}
 }\\
}

\date{}

\begin{document}

\maketitle


\input{secs/abstract}
\input{secs/introduction}

\input{secs/related_work}

\input{secs/method}

\input{secs/experiments}

\input{secs/conclusion}


\section{Limitations}
Our study has two main limitations. First, the final verification stage in the data construction pipeline relies on a strong VLM judge (Gemini-2.5-Pro), which may introduce model-dependent preferences in ambiguous cases despite the largely deterministic and rule-based earlier stages. Second, because MedGround is built from public segmentation datasets, it inherits their coverage limits in label space, imaging modality, and patient population, so some anatomical regions, pathologies, and acquisition settings remain underrepresented and performance may degrade under distribution shift.

\clearpage

\bibliographystyle{plainnat}

\bibliography{custom}

\appendix
\clearpage
\input{secs/Appendix}

\newpage
\input{checklist.tex}

\end{document}

%% file: secs/abstract.tex
\begin{abstract}
\textbf{V}ision-\textbf{L}anguage \textbf{M}odels (VLMs) can generate convincing clinical narratives, yet frequently struggle to visually ground their statements. We posit that this limitation arises from the scarcity of high-quality, large-scale clinical referring-localization pairs. To address this, we introduce MedGround, an automated pipeline that transforms segmentation resources into high-quality medical referring grounding data. Leveraging expert masks as spatial anchors, MedGround precisely derives localization targets, extracts shape and spatial cues, and guides VLMs to synthesize natural, clinically grounded queries that reflect morphology and location. To ensure data rigor, a multi-stage verification system integrates strict formatting checks, geometry- and medical-prior rules, and image-based visual judging to filter out ambiguous or visually unsupported samples. Finally, we present MedGround-35K, a novel multimodal medical dataset. Extensive experiments demonstrate that VLMs trained with MedGround-35K consistently achieve improved referring grounding performance, enhance multi-object semantic disambiguation, and exhibit strong generalization to unseen grounding settings. This work highlights MedGround as a scalable, data-driven approach to anchor medical language to verifiable visual evidence.
\end{abstract}

%% file: secs/introduction.tex
\begin{figure}
    \centering
    \includegraphics[width=1\linewidth]{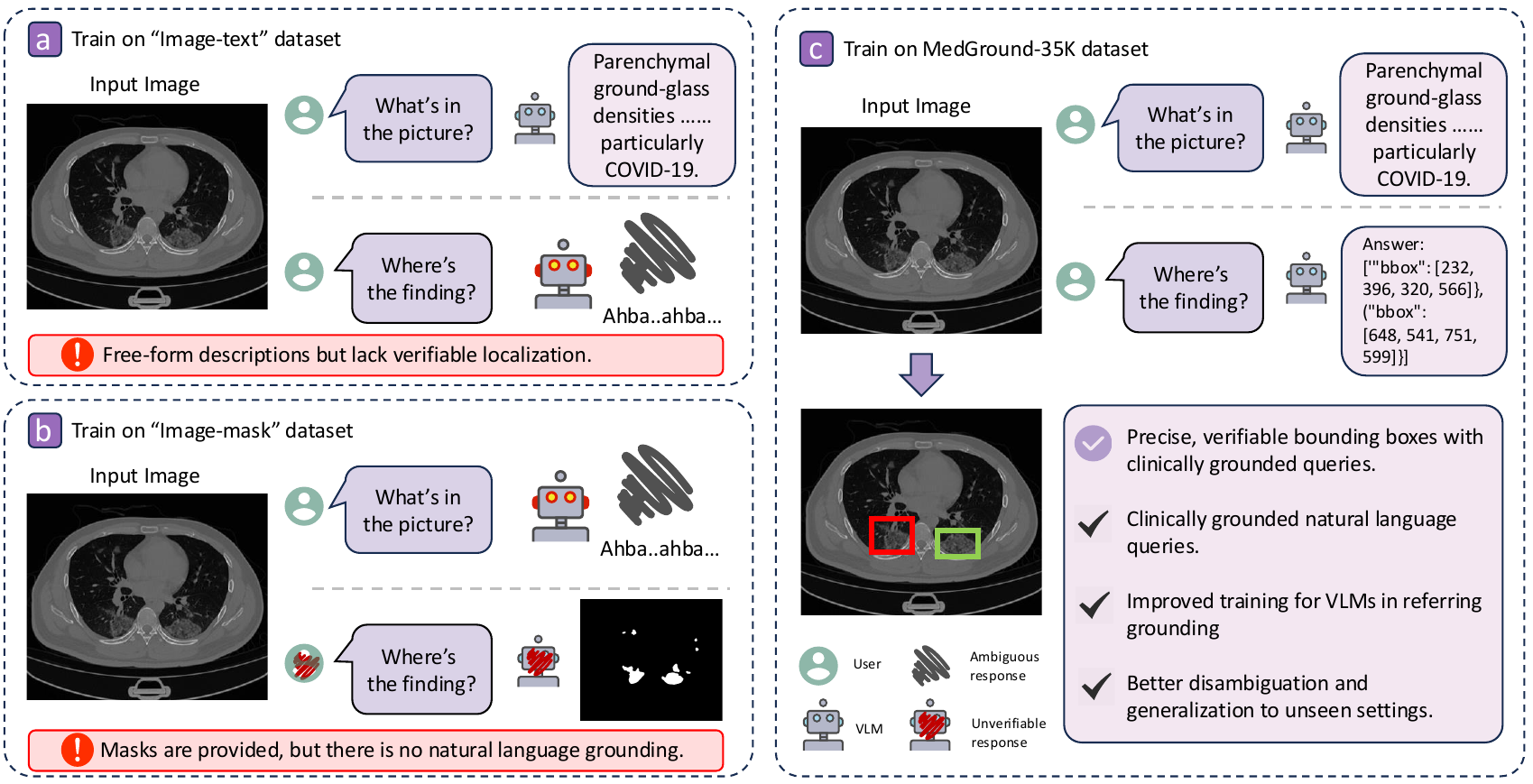}
    \caption{\textbf{Motivation of MedGround.} (a) Models trained on image-text pairs fail to \textit{"speak with substance"} due to lack of grounding. (b) Segmentation-only training fails to achieve semantic understanding. (c) \textbf{MedGround} (Image-text-box triplets) activates the full potential of medical VLMs by bridging semantics and localization.}
    \vspace{-3mm}
    \label{fig:motivation}
\end{figure}

\section{Introduction}
Recent VLMs have shown impressive capability in medical image understanding tasks such as report generation and clinical question answering \cite{li2023llava,manzari2023medvit,nath2025vila,sellergren2025medgemma,liu2025application, yang2024advancing, tarhini2025general,wu2025towards}. However, their linguistic fluency often outpaces fine-grained visual localization: a model may describe plausible findings while failing to identify where those findings appear in the image. Such visually unfaithful outputs undermine interpretability and can lead to “right-for-the-wrong-reason” predictions \cite{kim2025medclm,xie2023faithful,mahmood2025evaluating,ostmeier2024green,pal-etal-2023-med,huang2025towards,xu2025uncertainty}. 
We characterize this mismatch as a cognitive–perceptual gap, where a model’s cognitive competence is not mirrored by its perceptual grounding.

We argue that a key driver of this gap is the current data scarce. Medical data is plentiful in two largely disconnected forms: (1) image–text pairs from radiology reports, providing rich global semantics but weak spatial supervision; and (2) image–mask segmentation datasets, providing precise region annotations but impoverished language, often limited to a coarse category label. In contrast, large-scale datasets that pair natural referring expressions with explicit localization targets (boxes or masks) are scarce\cite{bannur2024maira,li2024anatomical,chen2025mimo,deng2025med} .This scarcity limits the ability of VLMs to learn the alignment between clinically meaningful phrases and localized evidence.

However, it is challenging to construct medical referring grounding data on a scale. One seemingly naive approach is to ground existing reports by asking VLMs to localize findings. Yet current VLMs are often vision-weak at fine-grained grounding, so this direction risks inheriting and amplifying their spatial biases\cite{huang2024refer,ge2025clinkd,strudel2022weakly,bai2024m3d,zhang2025generalized}. Instead, we reverse the process: starting from precise expert annotations in segmentation datasets, we use them as deterministic spatial anchors to synthesize referring expressions. This leverages the asymmetry of current models—strong language generation but weaker grounding—while maintaining spatial faithfulness through verification.

We introduce MedGround, a mask-guided semantic synthesis and verification pipeline that converts segmentation annotations into high-quality image–text–box triplets(MedGround-35K dataset) for medical referring expression grounding. MedGround effectively generates these triplets by leveraging expert masks to derive precise bounding boxes and extract geometric and spatial attributes. It then prompts a VLM to synthesize medically meaningful referring queries, followed by a multi-stage filtering process, including image-based VLM judging, to ensure data quality.  Extensive experiments across multiple settings show that training with our constructed  MedGround-35K dataset effectively improves medical referring grounding, and that the introduced clinically grounded semantics help VLMs better follow morphology- and location-aware descriptions for more reliable target disambiguation. These findings support MedGround as an effective and scalable supervision pipline for bridging the cognitive--perceptual gap. In summary, our key contributions are:
\begin{itemize}
    \item We propose MedGround, a scalable pipeline for synthesizing and verifying medically grounded referring queries anchored to expert annotations.
    \item We release MedGround-35K, covering eight datasets and multiple modalities, and demonstrate improvements in referring grounding, semantic disambiguation, and zero-shot transfer.
    \item We extensively tested various models using the MedGround-35K dataset and discovered that existing VLMs commonly struggle with fine-grained medical referring grounding tasks. However, training with our data significantly mitigates these issues.
\end{itemize}
\vspace{-5mm}

%% file: secs/related_work.tex
\newcommand{\cmark}{\textcolor{green!60!black}{\ding{51}}}
\newcommand{\xmark}{\textcolor{red!70!black}{\ding{55}}}
\newcommand{\tmark}{\textcolor{yellow!60!black}{$\triangle$}}

\section{Related Work}

\begin{table}
\centering
\caption{\textbf{Comparison of medical datasets.} \cmark: provided; \xmark: not provided; \tmark: limited/weak or implicit/derived. Auto Anno. means the samples are collected autonomously. }
\scriptsize
\setlength{\tabcolsep}{4pt}
\renewcommand{\arraystretch}{1.1}
\begin{adjustbox}{max width=\textwidth}
\begin{tabular}{lccccccl}
\hline
\textbf{Dataset} & \textbf{Modality} & \textbf{Text Type} & \textbf{Loc. Type} & \textbf{Granularity} & \textbf{Auto Anno.}  & \textbf{\#Anno.} & \textbf{Task} \\
\hline
MS-CXR\cite{boecking2022ms}& CXR & \cmark\ report phrases & \cmark\ box & finding & \xmark & 1,162 & phrase grounding \\
Chest ImaGenome \cite{wu2021chest}& CXR & \cmark\ structured phrases & \tmark\ region-aligned & region/finding & \xmark & 1,256 & region-text alignment \\
ChestX-ray8 \cite{chestxray14} & CXR & \tmark labels & \cmark box & disease & \xmark & 888 &disease localization \\
DeepLesion \cite{deeplesion_miccai2018}& CT & \xmark\ metadata & \cmark\ box & lesion & \xmark & 32,120 & lesion detection \\
LIDC-IDRI \cite{armato2011lidc}& CT & \xmark\ attributes & \cmark\ mask/contour & lesion & \xmark & 875 & nodule analysis \\
\hline
\textbf{MedGround-35K (ours)} & multi & \cmark\ referring queries & \cmark\ box & lesion/structure & \cmark & 35,480 & referring grounding \\
\hline
\end{tabular}
\end{adjustbox}
\end{table}

\begin{table}
    \centering
    \small
    \caption{\textbf{The statistics of the MedGround-35K.} The Anno. Tokens and Avg. Words columns show the total number of tokens and the average number of words for the medical grounding annotations regardless of task templates. The Modalities/Sources column shows the number of unique medical imaging sources/modalities involved in each split.}
    \label{tab:medground-stats}
    \resizebox{\textwidth}{!}{ 
    \begin{tabular}{lccccc}
    \toprule
    Split & \#Images & Anno. Tokens & Avg. Words &  Modality Ratio \\ 
    \midrule
    Train & 25.4k & 2773.5k & 12.0 &  Bacteria: 1.3\%, CT: 8.2\%, Dermoscopy: 9.9\%, Nuclei: 20.5\%, Ultrasound: 60.2\% \\ 
    \midrule
    Test  & 10.1k  & 1115.9k & 12.7 & Bacteria: 1.2\%, CT: 18.3\%, Dermoscopy: 11.8\%, Nuclei: 25.3\%, Ultrasound: 43.4\% \\ 
    \bottomrule
    \end{tabular}
    } 
\end{table}

\subsection{Medical Vision–Language Models}
Medical VLMs have rapidly advanced with large-scale pretraining on radiology report corpora and biomedical image–text resources, followed by instruction tuning for clinical question answering and report generation\cite{li2023llava,moor2023med,zhang2023pmc}. These models can produce fluent, clinically plausible narratives, but their outputs are not always evidence-aligned: models may describe findings without reliably localizing the corresponding visual regions, which limits interpretability and can lead to visually unfaithful reasoning\cite{pellegrini2023radialog,moll2025evaluating,liu2023qilin,gundersen2025enhancing}. This has motivated growing interest in grounding-aware evaluation and training signals that connect medical language to spatial evidence. In this work, we target this limitation by providing explicit referring-style localization supervision derived from expert segmentation annotations, enabling medical VLMs to better align morphology- and location-bearing phrases with concrete visual evidence.

\subsection{Referring Expression Grounding }
Referring expression grounding localizes an entity specified by a natural-language expression, emphasizing attribute understanding and spatial disambiguation. In natural images, large-scale referring and phrase grounding benchmarks have driven progress in models that bind text tokens to localized regions and handle fine-grained relational language\cite{liu2024grounding,li2022grounded,kamath2021mdetr}. In medical imaging, however, referring expression grounding datasets remain scarce. Available supervision is typically split between image--text pairs (rich clinical narratives but weak spatial alignment) and segmentation masks (precise localization but little to no language beyond class labels). This gap prevents medical VLMs from learning clinically meaningful referring cues such as morphology, laterality, anatomical sub-location, and multi-finding disambiguation\cite{bai2024m3d,deng2025med,liu2023qilin,zhang2025anatomical,yang2025new}. MedGround bridges this gap by converting segmentation annotations into scalable image–text–box supervision, explicitly training the model to follow clinically grounded referring descriptions.

\subsection{VLMs-based Synthesis and Verification}
VLMs-driven dataset construction has become a practical route to scale instruction and annotation resources, often paired with automated filtering, self-checking, or model-based judging to control noise. In medical settings, such synthesis is particularly fragile: hallucinated attributes, incorrect spatial relations, or visually unsupported statements can easily slip into training data and degrade evidence faithfulness\cite{pal-etal-2023-med,yu2023evaluating,khanna2023radgraph2,dong2023raft,croxford2025automating,yue2025medsg}. MedGround adopts a conservative synthesis principle: rather than asking a model to discover or localize findings from free-form text, we start from expert masks as deterministic spatial anchors and only synthesize referring queries conditioned on mask-derived geometry and spatial cues. We then apply multi-stage verification—format constraints, rule-based geometric/medical priors, and image-based judging—to filter ambiguous or visually unsupported samples. This design explicitly reduces the risk of inheriting vision-side localization bias while retaining the scalability benefits of VLMs-based generation.

%% file: secs/method.tex
\begin{figure*}
    \centering
    \includegraphics[width=1\linewidth]{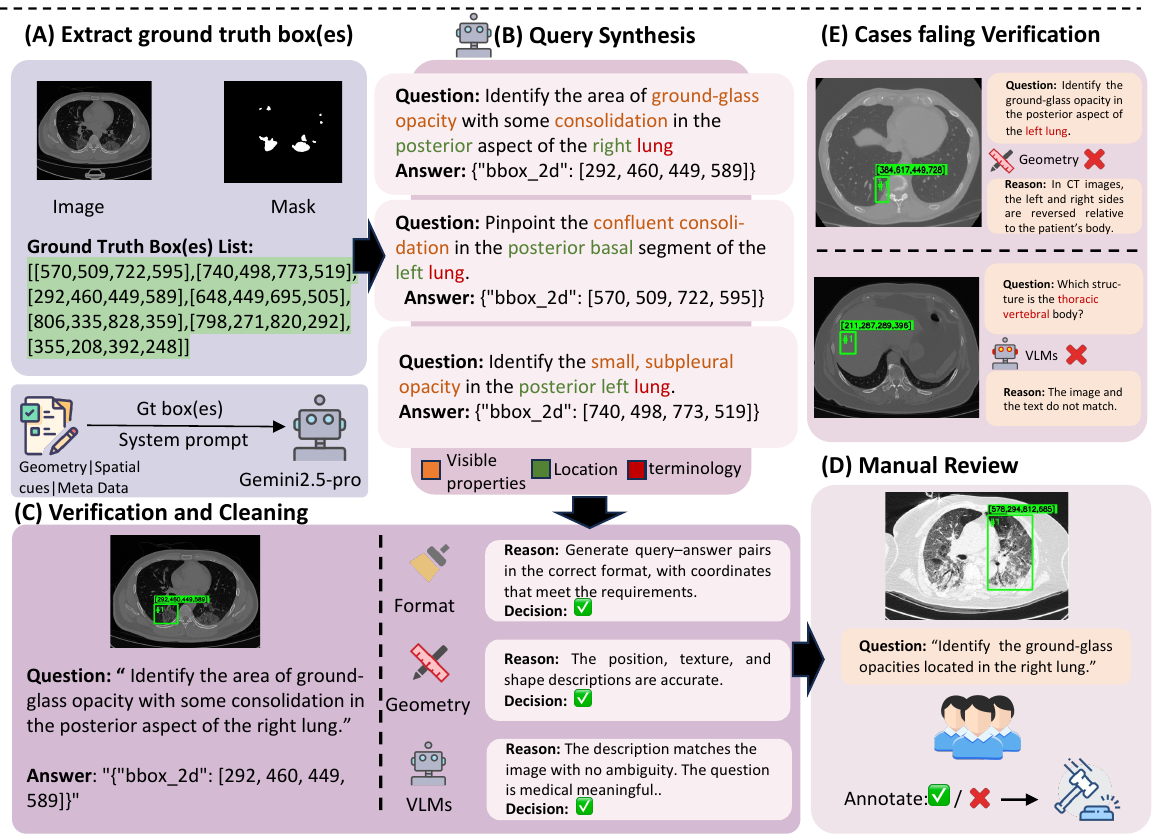}
    \caption{\textbf{MedGround pipeline.} (A) Convert segmentation masks into normalized ground-truth bounding box lists. (B) Use dataset-aware, mask-guided prompts to synthesize medically meaningful referring queries and select target box(es) as answers. (C) Perform multi-stage verification and cleaning (format/schema, geometry–location rules, and VLM-based grounding). (D) Conduct manual review for final quality control. (E) Cases falling verification.}
    \vspace{-3mm}
    \label{fig:pipline}
\end{figure*}

\section{MedGround}
This section introduces \textbf{MedGround}, a mask-guided synthesis-and-verification pipeline (Fig.~\ref{fig:pipline}) that automatically converts segmentation annotations into high-quality \emph{medical referring box grounding} datasets. 
Starting from expert masks, MedGround derives candidate ground-truth boxes, prompts a VLM to generate clinically grounded referring queries for randomly selected targets, and applies multi-stage verification to filter ambiguous or visually unsupported samples.

\subsection{Dataset Definition}
We produce MedGround-35K with MedGround pipline. Each example of MedGround-35K is a triplet $(I, Q, B)$, where $I$ is a medical image, $Q$ is a referring query that describes one or multiple target regions, and $B$ is the corresponding ground-truth set of 2D bounding boxes. Each box is represented as $[x_{\min}, y_{\min}, x_{\max}, y_{\max}]$ and normalized to a $1000 \times 1000$ coordinate grid for consistent formatting across datasets and resolutions.

\subsection{From Masks to Ground-Truth Box Lists}
MedGround-35K is built from eight public segmentation datasets(Tab.\ref{tab:data_source}), spanning dermatology, microscopy, and radiological imaging. Each dataset provides expert pixel masks $M$. For each connected component in $M$, we deterministically derive a tight bounding box $b=\mathrm{bbox}(M)$, and collect all boxes within the same image into a candidate list $\mathcal{B}=\{b_1,\ldots,b_n\}$.
This box list serves two roles: it provides the query generator (VLMs) with a \emph{candidate target pool}, i.e., explicit regions to refer to when composing queries, from which one or multiple boxes are randomly selected as the intended target(s); and it serves as the \emph{ground-truth localization anchor} during verification, supplying the reference boxes needed to judge whether the generated text is visually faithful and unambiguous with respect to the image.

\subsection{Mask-Guided Query Synthesis}
Given an image $I$ and its box list $\mathcal{B}$, we first compute mask-derived attributes to constrain generation:
\textbf{(1) geometry}: area ratio, width/height, aspect ratio, elongation/compactness proxies,
\textbf{(2) spatial cues}: centroid, coarse bins such as left/right and upper/lower, optionally with dataset-specific anatomical conventions,
and \textbf{(3) metadata}: imaging modality/domain and coarse category labels when available.

We then construct dataset-aware prompts and query VLMs to synthesize referring queries. The prompts explicitly condition on the image modality, and instruct the VLMs to produce questions that: 
(1) reflect \emph{visible} properties (shape/texture/boundary appearance), 
(2) incorporate location cues when needed for disambiguation, and 
(3) use appropriate medical terminology while avoiding claims that cannot be justified from the image (e.g., etiology, pathology stage, or non-visible symptoms).

To encourage diversity and avoid trivial templates, we ask the VLMs to \emph{randomly select} one or multiple target boxes from $\mathcal{B}$ and generate a corresponding query. Each generated sample is required to follow a fixed JSON schema containing the query text and the selected target box coordinates, enabling downstream automatic parsing.

\subsection{Multi-Stage Data Verification and Cleaning}
Since free-form generation may introduce ambiguity or visually unsupported descriptions, MedGround applies a multi-stage verifier to filter noisy samples.

\textbf{Stage \uppercase\expandafter{\romannumeral 1}: format and schema checks.}
We reject generations that do not conform to the required JSON format, contain missing fields, or output invalid box indices/coordinates.

\textbf{Stage \uppercase\expandafter{\romannumeral 2}: rule-based validation with geometry and location constraints.}
We enforce consistency between textual descriptors and mask-derived attributes. For example, size adjectives must match area buckets, spatial phrases must match centroid bins, and prompts are constrained by domain-specific keyword allow/deny lists to prevent cross-domain leakage (e.g., thoracic anatomy terms in dermoscopy). Samples violating these rules are discarded.

\textbf{Stage \uppercase\expandafter{\romannumeral 3}: VLM-based consistency filtering.}
We performs image-based data cleaning using a VLMs as a semantic verifier. Recall that during synthesis, the generator first selects one or multiple target box(es) from the candidate pool $\mathcal{B}$ and then writes a question to refer to the selected region(s); therefore, the ``answer'' $A$ of each QA pair is exactly the selected ground-truth box(es).

We feed the verifier the image together with the selected answer box(es) as an explicit localization anchor, and ask whether the visual content inside the box supports the question description. Concretely, the verifier is instructed to restate the observable attributes of the highlighted region and output a binary decision on whether the question is \emph{correctly grounded} and \emph{unambiguous}. If the verifier cannot recover the described cues from the given box, we treat the sample as hallucinated or ambiguous and discard it.

\begin{figure*}[t]
    \centering
    \includegraphics[width=1\linewidth]{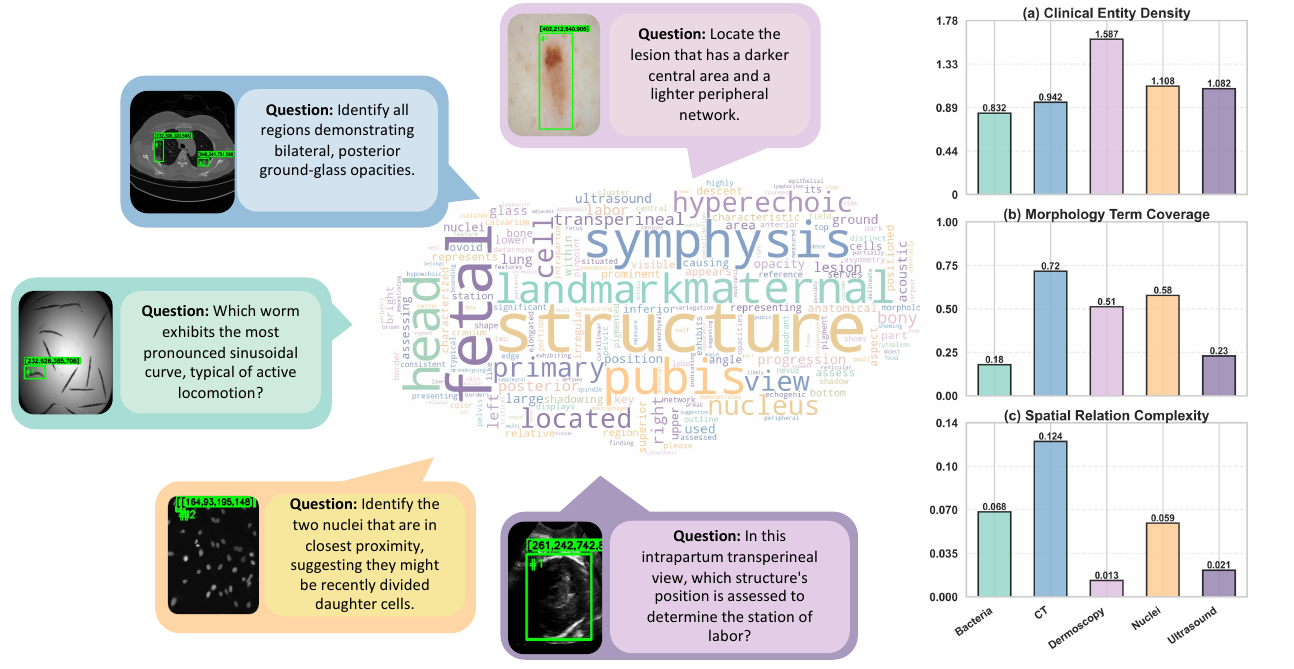}
    \caption{\textbf{Diversity and Linguistic Complexity of the MedGround dataset.} Up: The word cloud illustrates the distribution of medical terminology, anatomical landmarks, and clinical descriptors within the grounding annotations. Down: Comparative analysis of annotation richness across five distinct modalities based on three key metrics: (a) Clinical Entity Density, (b) Morphology Term Coverage, and (c) Spatial Relation Complexity.}
    \label{fig:distribution}
    \vspace{-5mm}
\end{figure*}

\subsection{MedGround-35K}
MedGround-35K comprises \textbf{$35,480$} image–text–box triplets, with \textbf{$25,420$} for training and $10,060$ for testing. To characterize the semantic richness of the synthesized queries, we quantify three linguistically grounded properties:\textbf{(1)Clinical entity density:} Measured by extracting unique UMLS\cite{bodenreider2004unified} concepts via SciSpacy and normalizing by query length. \textbf{(2)Morphology coverage:} Assessed using a modality-aware lexicon of appearance descriptors across CT, dermoscopy, microscopy, and ultrasound. \textbf{(3)Spatial complexity:} Calculated as the average frequency of spatial prepositions and relational phrases per query. Details semantic index of MedGround-35K are shown in Fig.\ref{fig:distribution}. In addition, we log the outcome of each verification stage and report the pass rate per stage in appendix(Tab.\ref{tab:stage_pass_rates}).

We also examine potential synthetic stylistic bias introduced by VLM-based query generation. Because human-written medical grounding expressions are scarce outside chest X-ray, we use radiology as a proxy setting and compare generated queries with MIMIC-CXR text. The appendix~\ref{tab:linguistic_bias} shows that generated queries are more noun-centric, longer, and less hedge-heavy than clinical reports, while still maintaining substantial lexical and structural diversity. This suggests a measurable but functional style shift: the generated language differs from authentic clinical writing, yet retains medically relevant grounding content.


\subsection{Human Audit}
To estimate the faithfulness of retained triplets and quantify residual noise, we conduct a full human audit on the entire MedGround-35K test set. Each triplet is independently reviewed by \textbf{three trained professional medical annotators} and marked as \textit{good} if the referring query is clinically sensible and the target localization is visually consistent. We use \textbf{majority vote} to define high-confidence samples: a triplet is considered accepted if it receives at least two \textit{good} votes ($\geq$2/3).

Overall, we conducted a human audit of MedGround-35K by sampling test triplets and computing the majority-vote acceptance rate, which is \textbf{78\%}. This suggests that the majority of synthesized queries are both faithful to the underlying content and visually grounded. The per-dataset audit results are reported in Appendix Tab.~\ref{tab:human_audit}.

During human auditing, most rejected samples were associated with ambiguous descriptions. In the supplementary material, we further analyze these failure cases and present representative examples in Fig.~\ref{fig:manual_fail_cases}. We observe that the majority of failures stem from non-unique descriptions in images containing multiple similar objects, a phenomenon that is particularly common in microorganism and cell/nucleus datasets such as BBBC010, CellNuclei, MoNuSAC, and DeepBacs. We regard such ambiguity as a form of \textbf{natural noise} that is difficult to eliminate entirely in real-world medical grounding scenarios. Because these low-agreement cases constitute only a small fraction of MedGround, we retain them in order to preserve both data coverage and realism. Their audit statistics and common error patterns are provided in the appendix.

\textbf{Note:} In the experiments below, we train all models on the \textbf{original training split} to better reflect realistic data conditions and avoid human-selection bias. We also performed a full cleaning of the training split using the same vote-based verification protocol applied to the test split, and this cleaned version will be released together with the dataset. For evaluation, however, we report results on the \textbf{human-verified test split}, which includes only triplets that pass human auditing, in order to ensure reliable and consistent measurement.

%% file: secs/experiments.tex
\definecolor{pospurple}{HTML}{7500AA} 
\definecolor{negpurple}{HTML}{F7B3AE} 

\begin{table*}[t]
\centering
\caption{\textbf{Medical referring grounding performance across benchmarks.} We compare base models and their MedGround-finetuned counterparts. Colored deltas indicate changes over the corresponding base model (gain: \textcolor{pospurple}{purple}, drop: \textcolor{negpurple}{pink}).}
\label{tab:main_results}
\scriptsize
\setlength{\tabcolsep}{2.5pt}
\renewcommand{\arraystretch}{1.1}
\begin{adjustbox}{max width=\textwidth}
\begin{tabular}{llc|cccccccc}
\toprule
\textbf{Type} & \textbf{Model} & \textbf{Size} &
\textbf{ISIC2016} & \textbf{BBBC010} & \textbf{BriFiSeg} & \textbf{CellNuclei} &
\textbf{DeepBACS} & \textbf{FHPsAOP} & \textbf{MoNuSAC} & \textbf{MosMedPlus} \\
\midrule
\multirow{7}{*}{General}
& Qwen2.5-VL-7B   & 7B & 9.9 & 0.3 & 1.1 & 0.5 & 0.0 & 0.1 & 2.3 & 1.3 \\
& Qwen3-VL-8B     & 8B & 81.7 & 44.7 & 13.2 & 15.9 & 9.0 & 21.0 & 13.6 & 11.1 \\
& InternVL3.5-4B  & 4B & 52.0 & 25.0 & 10.6 & 18.8 & 5.5 & 9.4 & 7.7 & 13.7 \\
& InternVL3.5-8B  & 8B & 53.4 & 28.2 & 14.8 & 26.6 & 10.0 & 7.3 & 9.0 & 16.8 \\
& MiniCPM-V-4.5   & 4B & 57.8 & 12.3 & 4.8 & 8.0 & 4.0 & 20.8 & 6.0 & 6.3 \\
& Qwen3.5-4B      & 4B & 54.7 & 17.4 & 6.8 & 4.5 & 0.7 & 11.0 & 5.4 & 6.7 \\
& Qwen3.5-9B      & 9B & 62.0 & 29.7 & 9.0 & 5.5 & 1.3 & 14.5 & 5.2 & 10.0 \\
\midrule
\multirow{7}{*}{\shortstack[l]{Medical\\VLMs}}
& Lingshu-7B            & 7B  & 54.8 & 6.1 & 5.0 & 3.2 & 2.7 & 15.4 & 6.0 & 6.6 \\
& MedGemma-27B          & 27B & 54.5 & 8.6 & 2.9 & 3.2 & 0.9 & 22.0 & 4.4 & 3.0 \\
& MedGemma-4B           & 4B  & 50.4 & 4.0 & 3.0 & 2.3 & 0.6 & 17.1 & 3.5 & 3.5 \\
& HuLuMed-4B            & 4B  & 6.3 & 1.2 & 1.1 & 0.5 & 0.0 & 1.8 & 1.3 & 0.7 \\
& HuLuMed-7B            & 7B  & 26.4 & 0.4 & 1.4 & 1.0 & 0.0 & 5.9 & 1.4 & 2.6 \\
& MedGemma1.5-4B-IT     & 4B  & 52.0 & 5.2 & 1.6 & 1.5 & 0.4 & 16.6 & 2.0 & 3.4 \\
& MedVLM-R1             & 4B  & 44.9 & 11.0 & 2.2 & 1.1 & 0.0 & 21.7 & 1.6 & 2.7 \\
\midrule
\multirow{14}{*}{Finetuned}
& Qwen2.5-VL-7B & 7B
& 83.0 {\textcolor{pospurple}{(+73.1)}} & 31.0 {\textcolor{pospurple}{(+30.7)}} & 20.2 {\textcolor{pospurple}{(+19.1)}} & 10.3 {\textcolor{pospurple}{(+9.8)}} & 7.1 {\textcolor{pospurple}{(+7.1)}} & 77.1 {\textcolor{pospurple}{(+77.0)}} & 10.9 {\textcolor{pospurple}{(+8.6)}} & 30.1 {\textcolor{pospurple}{(+28.8)}} \\
& Qwen3-VL-8B & 8B
& 86.4 {\textcolor{pospurple}{(+4.7)}} & 46.5 {\textcolor{pospurple}{(+1.8)}} & 24.6 {\textcolor{pospurple}{(+11.4)}} & 34.7 {\textcolor{pospurple}{(+18.8)}} & 23.0 {\textcolor{pospurple}{(+14.0)}} & 81.0 {\textcolor{pospurple}{(+60.0)}} & 13.5 {\textcolor{negpurple}{(-0.1)}} & 30.1 {\textcolor{pospurple}{(+19.0)}} \\
& InternVL3.5-8B & 8B
& 83.4 {\textcolor{pospurple}{(+30.0)}} & 34.9 {\textcolor{pospurple}{(+6.7)}} & 22.5 {\textcolor{pospurple}{(+7.7)}} & 38.4 {\textcolor{pospurple}{(+11.8)}} & 11.2 {\textcolor{pospurple}{(+1.2)}} & 75.7 {\textcolor{pospurple}{(+68.4)}} & 10.8 {\textcolor{pospurple}{(+1.8)}} & 32.9 {\textcolor{pospurple}{(+16.1)}} \\
& MiniCPM-V-4.5 & 4B
& 81.2 {\textcolor{pospurple}{(+23.4)}} & 28.1 {\textcolor{pospurple}{(+15.8)}} & 16.9 {\textcolor{pospurple}{(+12.1)}} & 19.1 {\textcolor{pospurple}{(+11.1)}} & 3.8 {\textcolor{negpurple}{(-0.2)}} & 71.5 {\textcolor{pospurple}{(+50.7)}} & 7.7 {\textcolor{pospurple}{(+1.7)}} & 23.8 {\textcolor{pospurple}{(+17.5)}} \\
& Qwen3.5-4B & 4B
& 81.8 {\textcolor{pospurple}{(+27.1)}} & 62.0 {\textcolor{pospurple}{(+44.6)}} & 30.7 {\textcolor{pospurple}{(+23.9)}} & 24.9 {\textcolor{pospurple}{(+20.4)}} & 14.7 {\textcolor{pospurple}{(+14.0)}} & 67.6 {\textcolor{pospurple}{(+56.6)}} & 13.9 {\textcolor{pospurple}{(+8.5)}} & 32.6 {\textcolor{pospurple}{(+25.9)}} \\
& Lingshu-7B & 7B
& 84.2 {\textcolor{pospurple}{(+29.4)}} & 44.1 {\textcolor{pospurple}{(+38.0)}} & 19.6 {\textcolor{pospurple}{(+14.6)}} & 9.7 {\textcolor{pospurple}{(+6.5)}} & 8.1 {\textcolor{pospurple}{(+5.4)}} & 79.0 {\textcolor{pospurple}{(+63.6)}} & 10.9 {\textcolor{pospurple}{(+4.9)}} & 36.3 {\textcolor{pospurple}{(+29.7)}} \\
& MedGemma-27B & 27B
& 81.2 {\textcolor{pospurple}{(+26.7)}} & 26.8 {\textcolor{pospurple}{(+18.2)}} & 11.6 {\textcolor{pospurple}{(+8.7)}} & 11.2 {\textcolor{pospurple}{(+8.0)}} & 0.4 {\textcolor{negpurple}{(-0.5)}} & 80.6 {\textcolor{pospurple}{(+58.6)}} & 9.0 {\textcolor{pospurple}{(+4.6)}} & 39.3 {\textcolor{pospurple}{(+36.3)}} \\
& MedGemma-4B & 4B
& 71.2 {\textcolor{pospurple}{(+20.8)}} & 16.3 {\textcolor{pospurple}{(+12.3)}} & 6.2 {\textcolor{pospurple}{(+3.2)}} & 4.7 {\textcolor{pospurple}{(+2.4)}} & 0.0 {\textcolor{negpurple}{(-0.6)}} & 72.8 {\textcolor{pospurple}{(+55.7)}} & 5.3 {\textcolor{pospurple}{(+1.8)}} & 33.1 {\textcolor{pospurple}{(+29.6)}} \\
& HuLuMed-4B & 4B
& 82.2 {\textcolor{pospurple}{(+75.9)}} & 24.9 {\textcolor{pospurple}{(+23.7)}} & 12.7 {\textcolor{pospurple}{(+11.6)}} & 11.4 {\textcolor{pospurple}{(+10.9)}} & 1.8 {\textcolor{pospurple}{(+1.8)}} & 65.2 {\textcolor{pospurple}{(+63.4)}} & 7.1 {\textcolor{pospurple}{(+5.8)}} & 28.9 {\textcolor{pospurple}{(+28.2)}} \\
& HuLuMed-7B & 7B
& 84.3 {\textcolor{pospurple}{(+57.9)}} & 39.5 {\textcolor{pospurple}{(+39.1)}} & 21.3 {\textcolor{pospurple}{(+19.9)}} & 19.4 {\textcolor{pospurple}{(+18.4)}} & 3.3 {\textcolor{pospurple}{(+3.3)}} & 73.7 {\textcolor{pospurple}{(+67.8)}} & 9.2 {\textcolor{pospurple}{(+7.8)}} & 36.2 {\textcolor{pospurple}{(+33.6)}} \\
& MedGemma1.5-4b-it & 4B
& 76.0 {\textcolor{pospurple}{(+24.0)}} & 14.5 {\textcolor{pospurple}{(+9.3)}} & 7.1 {\textcolor{pospurple}{(+5.5)}} & 6.1 {\textcolor{pospurple}{(+4.6)}} & 1.0 {\textcolor{pospurple}{(+0.6)}} & 71.9 {\textcolor{pospurple}{(+55.3)}} & 4.3 {\textcolor{pospurple}{(+2.3)}} & 20.3 {\textcolor{pospurple}{(+16.9)}} \\
& MedVLM-R1 & 4B
& 78.9 {\textcolor{pospurple}{(+34.0)}} & 30.8 {\textcolor{pospurple}{(+19.8)}} & 13.3 {\textcolor{pospurple}{(+11.1)}} & 6.5 {\textcolor{pospurple}{(+5.4)}} & 2.1 {\textcolor{pospurple}{(+2.1)}} & 64.5 {\textcolor{pospurple}{(+42.8)}} & 7.8 {\textcolor{pospurple}{(+6.2)}} & 16.0 {\textcolor{pospurple}{(+13.3)}} \\
\bottomrule
\end{tabular}
\end{adjustbox}
\vspace{-5mm}
\end{table*}

\section{Experiments}
In this section, we evaluate the effectiveness of \textbf{MedGround-35K} in enhancing the fine-grained grounding capabilities of VLMs. We focus on whether the synthesized image--text--box triplets can empower models with evidence-grounded reasoning and improve medical semantic alignment beyond generic label-based supervision.
\vspace{-3mm}

\subsection{Experiment Settings}
We evaluate models on three benchmarks that cover complementary aspects of medical grounding and generalization:

\textbf{Medical Referring Grounding}
We use the full test split of MedGround-35K to assess medical referring grounding performance. Each sample consists of an image, a referring query, and the corresponding target box, requiring the model to localize the region described by the query.

\begin{figure}
    \centering
    \includegraphics[width=0.9\linewidth]{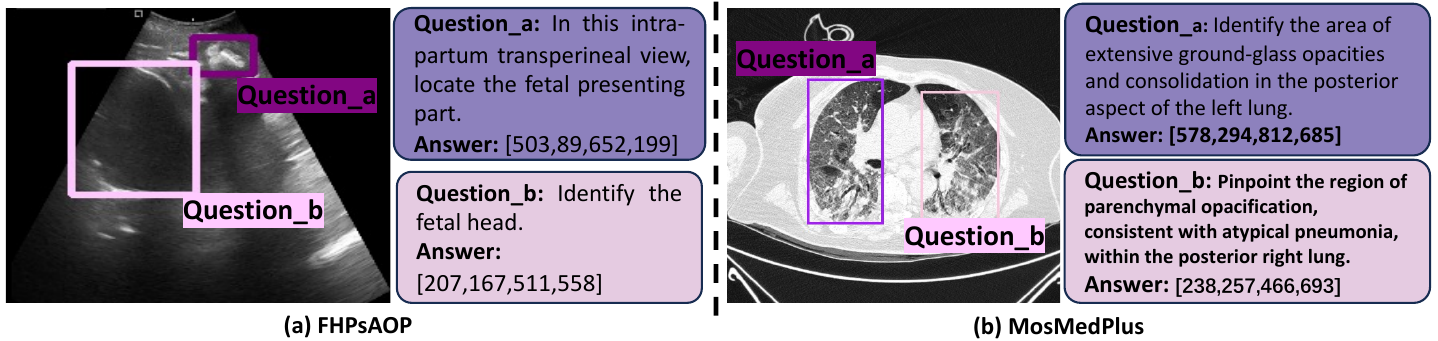}
    \caption{\textbf{Semantic dataset illustration.} The figure presents a paired semantic dataset constructed by describing two different objects within the same image, requiring the model to understand the semantics and associate them with the corresponding image content.}
    \vspace{-5mm}
    \label{fig:semantic}
\end{figure}

\textbf{Semantic Alignment}
To evaluate fine-grained semantic alignment, we build a semantic test set from MosMedPlus\cite{morozov2020mosmeddata} and FHPsAOP\cite{lu2022jnu,jieyun2024pubic}, both of which contain multiple targets per image. We select images with multiple annotated targets and form question pairs that refer to different targets within the same image (Fig.~\ref{fig:semantic}). This setup tests whether a model can resolve subtle semantic differences and localize the correct instance.

\textbf{Zero-shot Generalization}
We also evaluate zero-shot transfer on QaTa-COV19\cite{degerli2021covid}, without any task-specific fine-tuning on this dataset, to measure out-of-distribution generalization in medical grounding. This setting reflects how well the model transfers to unseen medical dataset.

\textbf{Evaluation Metrics}
We use IoU-based grounding accuracy as the primary metric. For semantic alignment, we further introduce \textbf{Semantic Sensitivity} (SS): a test case receives a score of 1 only if \emph{both} queried targets are correctly localized under an IoU threshold $\tau$, and 0 otherwise. We report SS as the average score over all test cases in the semantic benchmark.

\textbf{Training Details}
Training configurations and hyper-parameters are provided in Appendix B.1.

\subsection{Results and Analysis}

\paragraph{MedGround effectively enhances VLMs' medical referring grounding capability.} 
We evaluate the performance of VLM models trained on MedGround-35K. As shown in Tab.\ref{tab:main_results}, models fine-tuned with our knowledge-aware triplets demonstrate significant performance gains. The experimental results demonstrate that fine-tuning on the MedGround-35K serves as a transformative catalyst for the medical referring grounding capabilities of both general and medical-specific VLMs. By providing high-quality spatial-instructional alignment, MedGround enables models to overcome the limitations of zero-shot reasoning.
This consistent improvement across diverse modalities and model scales underscores MedGround’s efficacy in distilling specialized medical spatial knowledge into large-scale models, effectively bridging the gap between general visual perception and precise clinical localization.

While MedGround-35K significantly improved overall performance, the MedGemma series showed minor regressions on DeepBACS\cite{spahn2022deepbacs,spahn2021deepbacs} (e.g., -0.5 for MedGemma-27B). This is likely due to the domain shift between specialized fluorescence microscopy and the macro-level clinical imagery in MedGround. Fine-tuning may have caused a slight "feature forgetting" of niche microscopic traits while prioritizing general anatomical logic—a minor trade-off compared to the substantial gains achieved across the broader clinical spectrum.

\begin{figure}
    \centering
    \includegraphics[width=0.9\linewidth]{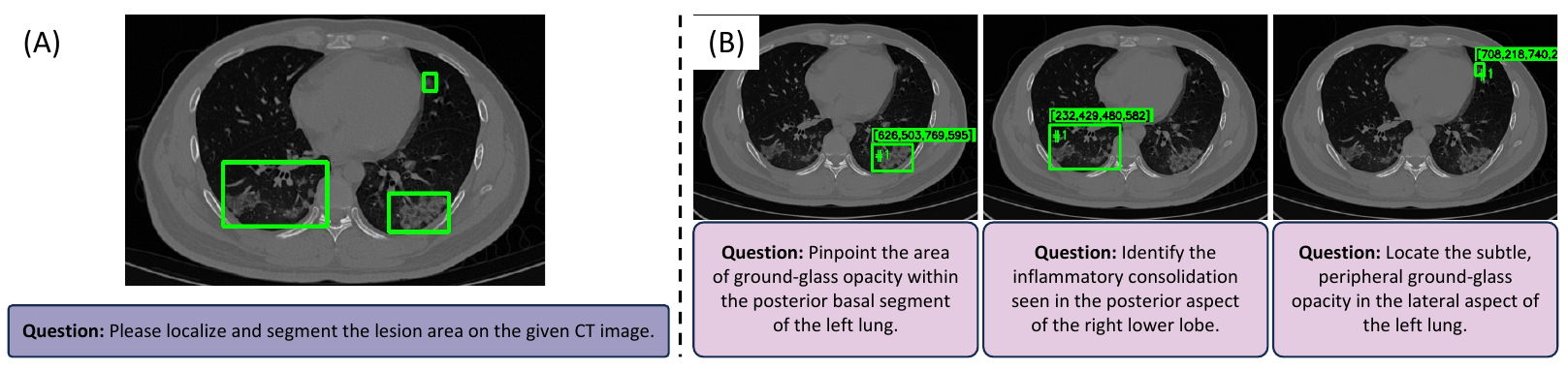}
    \caption{\textbf{Label-based Dataset vs. MedGround-35K.} (A) Label-based Dataset: conventional datasets typically group all ground-truth boxes under a single generic category label (e.g., "lesion") for the entire image. (B) MedGround-35K: provides distinct, fine-grained descriptive expressions for each localized box, capturing specific clinical nuances for individual regions.}
    \label{fig:compared-dataset}
\end{figure}

\begin{table}[t]
    \centering
    \caption{\textbf{Comparison of the SS metric.} Grounding performance is reported in Semantic Sensitivity (\%). Comparison across zero-shot baselines, label-based SFT, and MedGround SFT. For each cell, the best value among the three variants is shown in \textbf{bold}.}
    \label{tab:SS}
    \scriptsize
    \setlength{\tabcolsep}{1.6pt}
    \renewcommand{\arraystretch}{1.0}
    \resizebox{\columnwidth}{!}{%
    \begin{tabular}{llccccccccc}
    \toprule
    \textbf{Dataset} & \textbf{Variant} 
    & \textbf{Qwen2.5}
    & \textbf{Qwen3}
    & \textbf{Qwen3.5}
    & \textbf{InternVL3.5}
    & \textbf{MedGemma}
    & \textbf{MedGemma-1.5}
    & \textbf{MiniCPM-v4.5}
    & \textbf{HuLUMED}
    & \textbf{MedVLM-R1} \\
    
    & 
    & 7B
    & 8B
    & 4B
    & 8B
    & 4B/27B
    & 4B
    & -
    & 4B/7B
    & - \\
    \midrule

    \multirow{3}{*}{\textbf{FHPsAOP}}
    & w/o SFT 
    & 0.0 & 21.5 & 14.9 & 11.0 & 2.3 / 12.7 & 0.9 & 42.2 & 3.8 / 0.0 & 0.0 \\
    & Label-based SFT 
    & \textbf{71.9} & 45.3 & 51.3 & 15.2 & 21.3 / 27.0 & 18.6 & 52.3 & 37.7 / 51.0 & 12.7 \\
    & \textbf{MedGround-35K SFT} 
    & 74.2 & \textbf{53.5} & \textbf{80.4} & \textbf{79.2} & \textbf{55.9} / \textbf{74.3} & \textbf{73.4} & \textbf{78.2} & \textbf{44.5} / \textbf{81.7} & \textbf{26.8} \\
    
    \midrule

    \multirow{3}{*}{\textbf{MosMedPlus}}
    & w/o SFT 
    & 0.1 & 5.9 & 3.0 & 7.4 & 1.4 / 0.0 & 1.8 & 3.2 & 0.4 / 0.0 & 0.1 \\
    & Label-based SFT 
    & 15.6 & 5.8 & 17.3 & 6.2 & 4.4 / 2.3 & 5.3 & 7.2 & 14.9 / 16.8 & 3.2 \\
    & \textbf{MedGround-35K SFT} 
    & \textbf{16.4} & \textbf{18.0} & \textbf{23.9} & \textbf{24.6} & \textbf{13.1} / \textbf{22.5} & \textbf{8.7} & \textbf{12.5} & \textbf{22.5} / \textbf{19.4} & \textbf{6.5} \\
    \bottomrule
    \end{tabular}%
    }
\end{table}

\paragraph{MedGround injects fine-grained medical semantic knowledge into VLMs.}
We evaluate whether MedGround injects finer-grained clinical semantics than coarse label-level supervision using a Semantic Alignment setting. As shown in Tab.~\ref{tab:SS}, models without medical referring grounding training (\textit{w/o SFT}) exhibit very low Semantic Sensitivity across both benchmarks, indicating limited ability to disambiguate query-specific semantics and localize the intended targets in multi-object clinical images. After fine-tuning, MedGround-35K SFT achieves the strongest SS performance for the majority of model variants, and in most cases clearly outperforms the label-based SFT baseline, especially on FHPsAOP. This trend suggests that detailed clinical referring expressions provide supervision beyond coarse category names, which are often insufficient for distinguishing co-existing findings with different morphology, extent, or anatomical location.



\begin{figure}
    \centering
    \includegraphics[width=1.0\linewidth]{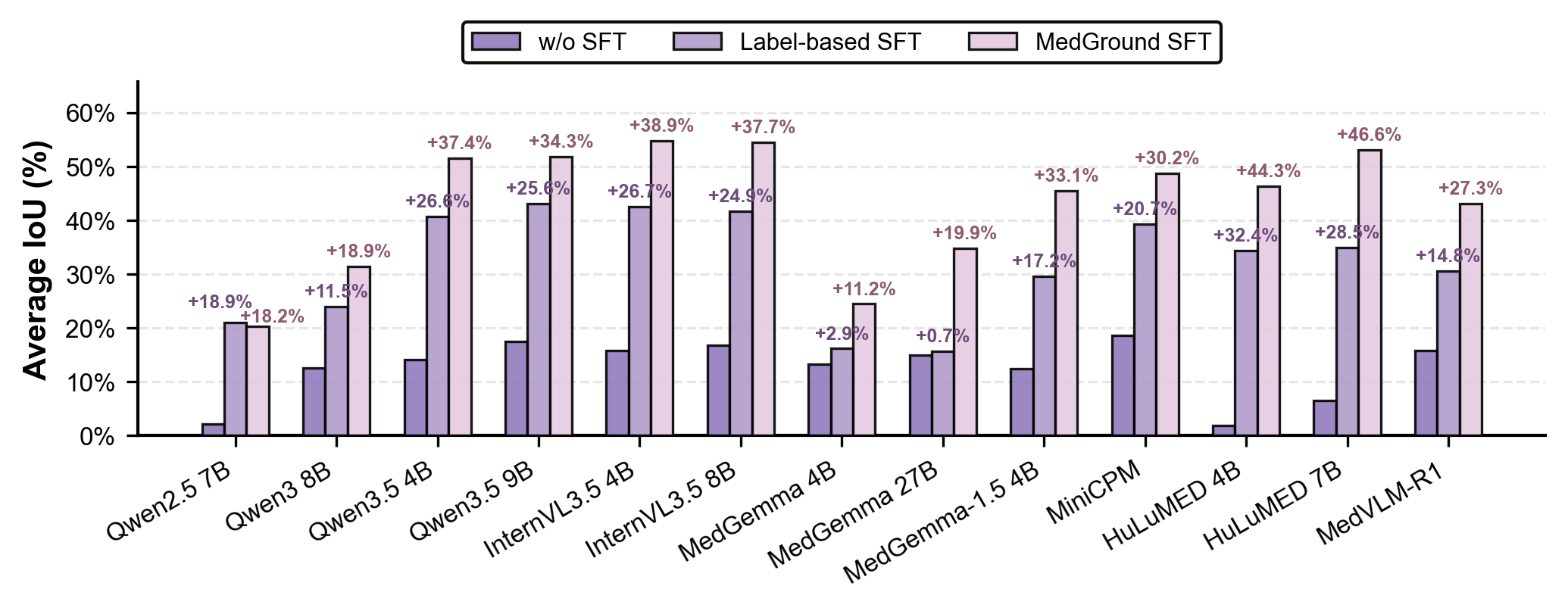}
    \caption{\textbf{Zero-shot Performance.} We evaluate different models on QaTa-COV19 dataset.}
     \vspace{-5mm}
    \label{fig:zero-shot}
   
\end{figure}

\paragraph{MedGround Enables Zero-shot Generalization of VLMs.}
We further evaluate whether MedGround improves cross-dataset transfer by testing on an external unseen dataset, QaTa-COV19. We construct a fine-grained referring grounding test set from its referring statements and ground-truth boxes, and compare the performance of the MedGround-fine-tuned model against the base model on this benchmark. The zero-shot evaluation on the external QaTa-COV19 dataset reveals that fine-tuning on MedGround-35K yields substantial and consistent improvements in cross-dataset transferability across all tested architectures. As illustrated in the Fig.\ref{fig:zero-shot}, the MedGround-35K enhanced models achieve remarkable performance gains compared to others. This significant performance leap on an entirely unseen dataset—achieved without any task-specific training—strongly demonstrates that MedGround does not merely facilitate dataset-specific memorization but rather imparts a robust, generalized medical spatial reasoning logic. Such results solidify MedGround's role as a critical foundational resource for enabling VLMs to generalize their referring grounding capabilities to novel clinical scenarios and emerging disease distributions.

%% file: secs/conclusion.tex
\section{Conclusion}
We introduce \textbf{MedGround}, a mask-guided synthesis and verification pipeline that constructs medical referring grounding data as image--query--box triplets. By generating fine-grained referring expressions and filtering ambiguous or visually unsupported samples, MedGround provides semantically rich, visually verified supervision. Fine-tuning on MedGround-35K consistently improves grounding performance and transfer to unseen benchmarks, helping narrow the \emph{cognitive--perceptual gap} by anchoring medical language to localized visual evidence. Since MedGround-35K is synthesized and verified from existing segmentation resources, it is highly scalable, enabling rapid extension to new modalities and anatomical systems. Beyond grounding, MedGround encourages models to justify clinical semantics with explicit spatial evidence, reducing reliance on linguistic priors and improving faithfulness.

%% file: secs/Appendix.tex
\label{sec:appendix}
\section{Datasets}
\subsection{Data source}

MedGround-35K is constructed by repurposing annotations from eight publicly available medical segmentation datasets, covering diverse imaging modalities and clinical scenarios, including BBBC010\cite{ljosa2012annotated}, BriFiSeg\cite{mathieu2022brifiseg}, CellNuclei\cite{caicedo2019nucleus}, DeepBacs\cite{spahn2022deepbacs}, FHPsAOP\cite{lu2022jnu}, ISIC2016\cite{codella2018skin}, MoNuSAC\cite{verma2021monusac2020}, and MosMedPlus\cite{morozov2020mosmeddata}. For each source dataset, we use its original train/val/test split as the raw data pool before applying our MedGround construction and verification pipeline. Detailed split statistics of these sources are summarized in Tab.~\ref{tab:data_source}.

\begin{table}[thbp]
\centering
\small
\setlength{\tabcolsep}{4pt}
\renewcommand{\arraystretch}{1.05}
\caption{Raw split sizes of segmentation data sources before MedGround construction.}
\label{tab:data_source}
\begin{tabular}{l r r r r}
\toprule
\textbf{Dataset} & \textbf{Train} & \textbf{Val} & \textbf{Test} & \textbf{Total} \\
\midrule
BBBC010        & 70     & 10    & 20    & 100 \\
Brifiseg       & 38{,}463 & 4{,}244 & 576   & 43{,}283 \\
CellNuclei     & 469    & 67    & 134   & 670 \\
DeepBacs       & 17     & 2     & 15    & 34 \\
FHPsAOP        & 5{,}600 & 800   & 1{,}600 & 8{,}000 \\
ISIC2016       & 810    & 90    & 379   & 1{,}279 \\
MoNuSAC        & 359    & 35    & 249   & 643 \\
MosMedPlus     & 1{,}910 & 271    & 547   & 2{,}457 \\
\midrule
\textbf{Total} & 47{,}689 & 5{,}529 & 3{,}520 & 56{,}466 \\
\bottomrule
\end{tabular}
\end{table}

\subsection{Pass rate of multi-stage verifaction pipline}
We analyze rejection reasons across stages (\uppercase\expandafter{\romannumeral 1}: format, \uppercase\expandafter{\romannumeral 2}: rule-based checks, \uppercase\expandafter{\romannumeral 3}: VLM judging) in Tab.~\ref{tab:stage_pass_rates}. Stages \uppercase\expandafter{\romannumeral 1} and \uppercase\expandafter{\romannumeral 2} retain almost all samples across datasets ($\sim$ 92--100\%), indicating that invalid formatting and obvious rule violations account for only a small fraction of failures. In contrast, Stage \uppercase\expandafter{\romannumeral 3} is substantially more selective and becomes the main source of filtering: retention drops to 48--67\% for several microscopy datasets (e.g., BBBC010, CellNuclei, DeepBACS, BriFiSeg), while remaining high for more visually distinctive settings such as FHPSAOP and ISIC2016 ($\sim$88--90\%). Overall, the final retention is about 80\% on both splits (80.9\% train, 80.3\% test), suggesting that the VLM judge primarily removes samples that are \emph{ambiguous} (multiple plausible referents) or \emph{visually unsupported} (the description does not clearly match the highlighted target). This aligns with our human audit, where most rejected cases are due to ambiguity, and supports that MedGround’s verification prioritizes unambiguous, evidence-grounded referring expressions rather than merely enforcing formatting constraints.

\begin{table*}[h]
\centering
\caption{\textbf{Pass rates of each verification stage across datasets.}  We report the retention percentage and remaining sample count after Stages A/B/C for both the training and test splits.}
\label{tab:stage_pass_rates}
\scriptsize
\setlength{\tabcolsep}{2pt} 
\renewcommand{\arraystretch}{1.2}

\begin{adjustbox}{max width=\textwidth}
\begin{tabular}{l|cc|cc|cc|cc|cc|cc|cc|cc|cc}
\toprule
\textbf{Stage} &
\multicolumn{2}{c|}{\textbf{BBBC010}} &
\multicolumn{2}{c|}{\textbf{BriFiSeg}} &
\multicolumn{2}{c|}{\textbf{CellNuclei}} &
\multicolumn{2}{c|}{\textbf{DeepBACS}} &
\multicolumn{2}{c|}{\textbf{FHPSAOP}} &
\multicolumn{2}{c|}{\textbf{ISIC2016}} &
\multicolumn{2}{c|}{\textbf{MoNuSAC}} &
\multicolumn{2}{c|}{\textbf{MosMedPlus}} &
\multicolumn{2}{c}{\textbf{Total}} \\
\cmidrule(lr){2-3}\cmidrule(lr){4-5}\cmidrule(lr){6-7}\cmidrule(lr){8-9}
\cmidrule(lr){10-11}\cmidrule(lr){12-13}\cmidrule(lr){14-15}\cmidrule(lr){16-17}\cmidrule(lr){18-19}
& \textbf{Train} & \textbf{Test}
& \textbf{Train} & \textbf{Test}
& \textbf{Train} & \textbf{Test}
& \textbf{Train} & \textbf{Test}
& \textbf{Train} & \textbf{Test}
& \textbf{Train} & \textbf{Test}
& \textbf{Train} & \textbf{Test}
& \textbf{Train} & \textbf{Test}
& \textbf{Train} & \textbf{Test} \\
\midrule
Stage \uppercase\expandafter{\romannumeral 1}  & 97.2\% & 97.0\% & 98.3\% & 98.4\% & 98.1\% & 98.4\% & 98.2\% & 97.2\% & 100.0\% & 100.0\% & 92.1\% & 92.4\% & 96.2\% & 96.7\% & 96.4\% & 94.8\% & 98.3\% & 97.6\% \\
Stage \uppercase\expandafter{\romannumeral 2} & 96.8\% & 97.0\% & 98.1\% & 98.0\% & 97.9\% & 98.3\% & 98.2\% & 97.2\% & 99.4\% & 99.2\% & 92.0\% & 92.3\% & 95.4\% & 95.8\% & 95.6\% & 94.5\% & 97.8\% & 97.1\% \\
Stage \uppercase\expandafter{\romannumeral 3}  & 56.9\% & 49.3\% & 67.1\% & 65.9\% & 57.1\% & 56.0\% & 48.2\% & 47.2\% & 89.7\% & 89.3\% & 88.2\% & 88.5\% & 63.8\% & 60.6\% & 78.4\% & 94.5\% & 80.9\% & 80.3\% \\
\midrule
\#Remaining & 265 & 66 & 2369 & 1212 & 1651 & 479 & 54 & 51 & 15291 & 4363 & 2523 & 1191 & 1193 & 855 & 2074 & 1843 & 25420 & 10060 \\
\bottomrule
\end{tabular}
\end{adjustbox}
\end{table*}

\subsection{Linguistic analysis of synthetic stylistic bias}
Because the queries in MedGround are generated by a VLM under a predefined prompt framework, an important question is whether they exhibit systematic synthetic stylistic bias. A direct modality-matched comparison against human-written medical grounding expressions is currently infeasible, since such annotations are largely unavailable for most medical modalities. We therefore perform a proxy analysis in radiology, where we compare generated chest X-ray queries from MosMedPlus/FHPsOAP against human-written text from MIMIC-CXR, using 500 samples from each group.

We evaluate lexical, syntactic, and stylistic properties, including type-token ratio, noun and verb density, sentence length, dependency depth, hedging frequency, readability, and trigram concentration.

\begin{table*}
\centering
\caption{Linguistic comparison between VLM-generated referring expressions and human-written radiology text.}
\label{tab:linguistic_bias}
\resizebox{\textwidth}{!}{%
\begin{tabular}{llcccc}
\toprule
Category & Metric & VLM (Mean$\pm$SD) & Human (Mean$\pm$SD) & p-value & Cohen's $d$ \\
\midrule
Lexical   & Type-Token Ratio     & 0.90$\pm$0.07 & 0.81$\pm$0.08 & $<.001$ & 1.215 \\
          & Noun Density         & 0.37$\pm$0.07 & 0.29$\pm$0.05 & $<.001$ & 1.301 \\
          & Verb Density         & 0.10$\pm$0.06 & 0.17$\pm$0.05 & $<.001$ & -1.362 \\
\midrule
Syntactic & Avg. Sentence Length & 13.56$\pm$3.06 & 9.56$\pm$2.91 & $<.001$ & 1.341 \\
          & Max Dependency Depth & 6.21$\pm$1.87 & 5.72$\pm$1.93 & $<.001$ & 0.259 \\
\midrule
Style     & Hedging Expressions  & 0.02$\pm$0.13 & 0.07$\pm$0.13 & $<.001$ & -0.411 \\
          & Flesch--Kincaid      & 13.46$\pm$3.52 & 10.38$\pm$2.03 & $<.001$ & 1.073 \\
\midrule
Variety   & Trigram Concentration & \textbf{0.2524} & 0.0498 & -- & -- \\
\bottomrule
\end{tabular}%
}
\end{table*}

The analysis reveals clear distributional differences between generated and human-written text. Compared with clinical reports, VLM-generated expressions are more noun-centric, contain fewer verbs and hedging expressions, and tend to be longer and stylistically more structured. These differences are expected because the generated outputs are optimized as discriminative referring expressions, whereas MIMIC-CXR reports are written for diagnostic communication.

At the same time, the generated expressions do not collapse into a narrow set of repetitive templates. Their lexical diversity remains high, and the phrase-level repetition statistics suggest nontrivial structural variation. This indicates that the stylistic bias is not merely a templating artifact, but rather a task-driven shift toward localization-oriented language.

Overall, the results suggest that MedGround exhibits a measurable but functional style gap relative to authentic clinical writing. We expect this gap to mainly affect linguistic style rather than medically relevant grounding content. Nevertheless, narrowing this gap with future collections of human-authored medical referring expressions remains an important direction for improving real-world generalization.

\subsection{Human Audit Results and Failure Analysis}
From the overall statistics, the lower pass rates are mainly observed in the \textbf{nuclei segmentation} datasets. By analyzing the auditors’ rejection reasons, we find that most failures stem from "ambiguous matching"---the referring description does not admit a unique target in the image, leading auditors to judge the image--text pair as not well aligned. This is consistent with the inherent properties of nuclei images, where scenes contain numerous visually similar instances and the language often relies on low-dimensional attributes that are insufficient for unique identification.
\begin{table*}[thpb]
\centering
\caption{\textbf{Human audit on the MedGround test split.} Each sample is reviewed by three auditors. We report the distribution of \textit{good} votes and the majority-pass rate (Good Ratio, $\geq$2/3 \textit{good}).}
\label{tab:audit_votes}
\small
\setlength{\tabcolsep}{3pt}
\renewcommand{\arraystretch}{1.15}

\begin{tabularx}{\linewidth}{@{}X r r r r r r@{}}
\toprule
\textbf{Dataset} & \textbf{Total} & \textbf{3 good} & \textbf{2 good} & \textbf{1 good} & \textbf{0 good} & \textbf{Good Ratio} \\
\midrule
ISIC2016\_512\_test     & 1,141 & 938 (82.2\%)  & 142 (12.4\%) & 34 (3.0\%)   & 27 (2.4\%)   & 94.65\% \\
mosmedplus\_512\_test   & 1,843 & 1,531 (83.1\%)& 57 (3.1\%)   & 49 (2.7\%)   & 206 (11.2\%) & 86.16\% \\
fhpsaop\_256\_test      & 4,363 & 3,167 (72.6\%)& 479 (11.0\%) & 233 (5.3\%)  & 484 (11.1\%) & 83.61\% \\
brifiseg\_512\_test     & 1,212 & 334 (27.6\%)  & 500 (41.3\%) & 283 (23.3\%) & 95 (7.8\%)   & 68.81\% \\
bbbc010\_512\_test      & 66    & 34 (51.5\%)   & 10 (15.2\%)  & 13 (19.7\%)  & 9 (13.6\%)   & 66.67\% \\
cellnuclei\_256\_test   & 479   & 88 (18.4\%)   & 147 (30.7\%) & 158 (33.0\%) & 86 (18.0\%)  & 49.06\% \\
monusac\_512\_test      & 855   & 129 (15.1\%)  & 267 (31.2\%) & 316 (37.0\%) & 143 (16.7\%) & 46.32\% \\
deepbacs\_512\_test     & 51    & 8 (15.7\%)    & 14 (27.5\%)  & 20 (39.2\%)  & 9 (17.6\%)   & 43.14\% \\
\bottomrule
\end{tabularx}
\label{tab:human_audit}
\end{table*}

\subsection{Bias analysis of the VLM judge}
Since our pipeline relies on a VLM as an automatic judge, we further analyze its potential biases. We identify a recurring failure mode in chest X-ray verification, where the judge may misinterpret anatomical laterality unless explicitly instructed to follow the radiological viewing convention. Under standard radiological display, the patient’s right side appears on the left side of the image. Without this prior, the judge may incorrectly reject otherwise valid triplets involving left/right references.

To mitigate this issue, we adopt a context-aware prompting strategy that explicitly reminds the judge of the radiological convention when evaluating chest X-ray samples. Table~\ref{tab:judge_bias} shows that this targeted prompting substantially improves verification performance on laterality-sensitive cases.

\begin{table}[t]
\centering
\caption{Effect of radiological-convention prompting on VLM judge verification for laterality-sensitive chest X-ray samples.}
\label{tab:judge_bias}
\scriptsize
\resizebox{\columnwidth}{!}{%
\begin{tabular}{lcccc}
\toprule
Metric & With Convention & Without Convention & Absolute $\Delta$ & Relative Improvement \\
\midrule
Overall Pass Rate & \textbf{97.22} & 88.95 & +8.27 & \textbf{+9.3\%} \\
Left Pass Rate    & 96.18 & 86.87 & +9.31 & +10.7\% \\
Right Pass Rate   & 97.97 & 90.40 & +7.57 & +8.4\% \\
\bottomrule
\end{tabular}%
}
\end{table}

The overall pass rate increases from 88.95\% to 97.22\%, with consistent improvements for both left- and right-sided references. This result suggests that the judge bias is partly systematic and can be effectively reduced through domain-specific prompt design.

More broadly, this analysis highlights a general limitation of VLM-based automatic filtering: verification quality may depend on implicit visual conventions that are not reliably captured unless explicitly encoded in the prompt. We therefore complement automatic filtering with human verification on the test split and release the human-verified benchmark for more reliable evaluation. We did not explore judge ensembles in the current version, but this remains a promising direction for improving robustness in future work.

\subsection{Train--Test Distribution Analysis in Three Semantic Dimensions}
\begin{figure*}[thpb]
    \centering
    \includegraphics[width=1\linewidth]{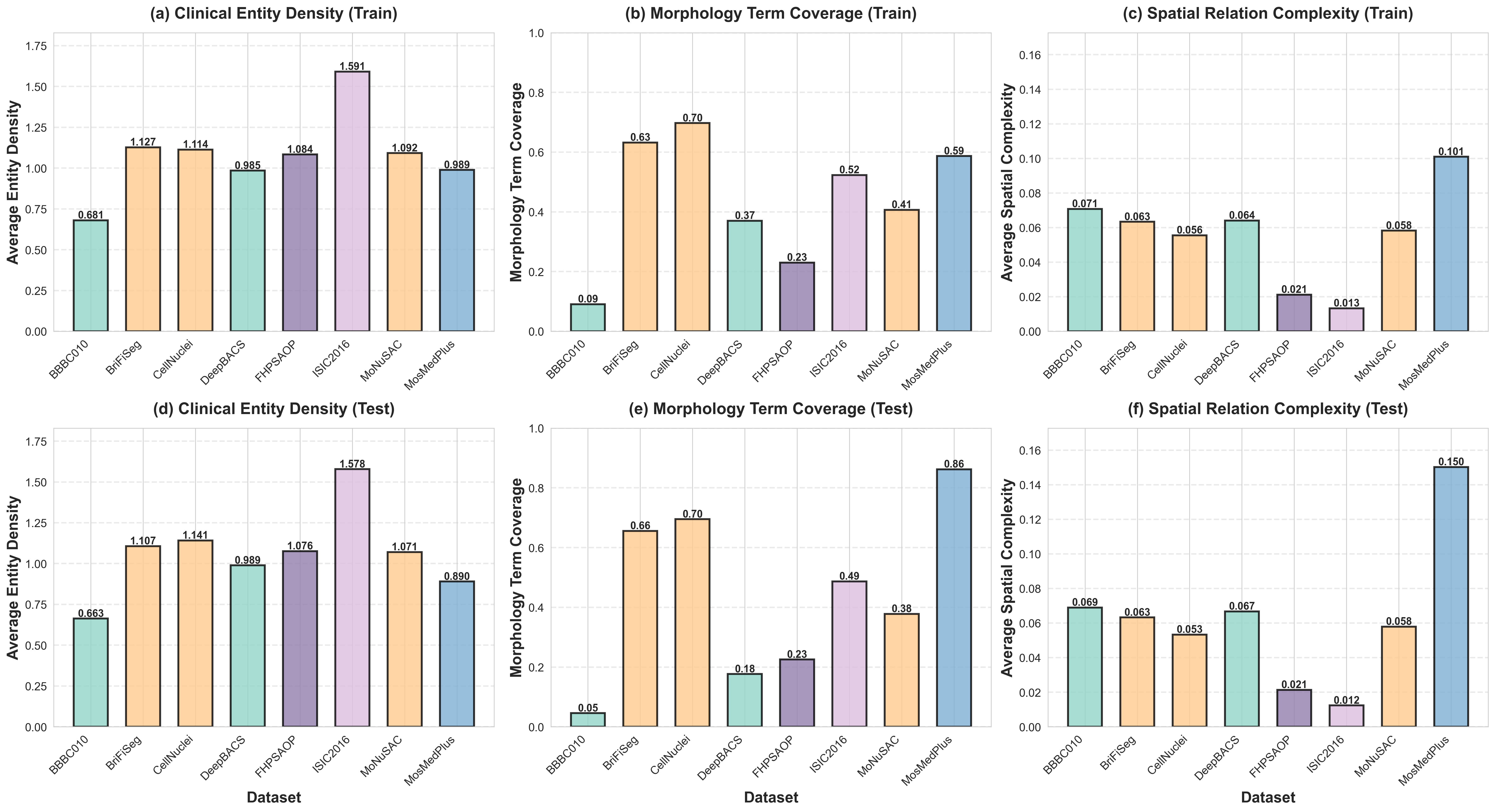}
    \caption{ Three-dimensional analysis comparing the training split (top row, a--c) and test split (bottom row, d--f) across eight medical imaging datasets. Colors: blue=CT, dark purple=ultrasound, light purple=dermoscopy, green=nuclei, orange=bacteria.}
    \label{fig:three-dimension}
\end{figure*}

 We quantify linguistic properties relevant to grounding(Fig.~\ref{fig:three-dimension}):
Clinical entity density using medical entity recognition/linking tools (e.g., UMLS-based pipelines).
Morphology term coverage using a curated lexicon of appearance descriptors.
Spatial relation complexity via counts of spatial prepositions and relational phrases.
These analyses show that MedGround queries contain substantially richer morphology and spatial language than category-only prompts, providing more informative supervision for grounding.

First, clinical entity density remains stable between splits for most datasets, indicating similar average numbers of clinically relevant referents per sample. This stability is important for evaluating grounding in multi-target scenes, since the number of co-existing entities directly affects ambiguity and search difficulty.

Second, morphology term coverage shows noticeable cross-dataset variation and a mild train–test gap for some sources. In particular, datasets such as MosMedPlus exhibit higher coverage in the test split, implying that the audited test set contains richer morphological descriptions and thus places greater emphasis on fine-grained visual-semantic alignment. Conversely, nuclei-related datasets (e.g., BBBC010, DeepBACS, MoNuSAC) generally present lower morphology coverage, reflecting the inherent limitation that many instances are visually similar and hard to differentiate with morphology alone.

Third, spatial relation complexity is consistently highest for MosMedPlus in both splits, reflecting that CT findings often require more location-aware language (e.g., lobes, peripheral vs. central, bilateral distribution) to uniquely specify targets. In contrast, pathology and dermoscopy datasets tend to have lower spatial complexity, consistent with their relatively localized lesions or simpler spatial context. This three-dimensional analysis highlights that different modalities stress different aspects of grounding: microscopy emphasizes dense-instance disambiguation, while CT emphasizes spatial reasoning, and the test split preserves (or slightly strengthens) these challenges.

\section{Implementation Details}
\subsection{Training Details}
We select MedGemma-27B, MedGemma-4B\cite{sellergren2025medgemma}, medgemma-1.5-4b-it\cite{sellergren2026medgemma15technicalreport}, Qwen2.5-VL-7B\cite{bai2025qwen2}, Qwen3-VL-8B\cite{bai2025qwen3vltechnicalreport}, Qwen3.5-4B/9B\cite{qwenteam2026qwen35omnitechnicalreport}, InternVL3.5-4B/8B\cite{wang2025internvl35advancingopensourcemultimodal}, hulumed-4/7B\cite{jiang2025hulumedtransparentgeneralistmodel}, medvlm-r1\cite{pan2025medvlmr1incentivizingmedicalreasoning} and minicmp\cite{hu2024minicpmunveilingpotentialsmall} as the base models and fine-tune them on the MedGround-35K training data to investigate how fine-grained clinical semantics enhance their medical referring grounding capabilities. To rigorously isolate the impact of linguistic granularity, we also fine-tune these models on a label-based baseline where detailed referring expressions are replaced by coarse category names. (Further implementation details and hyper-parameters are provided in appendix Tab.\ref{tab:hparams_qwen25vl7b}$\sim$\ref{tab:hparams_qwen3vl8b}).

\textbf{VLMs Evaluated} We compare with both general purpose VLMs and medical expert VLMs. During the evaluation, we manually craft grounding prompts suitable for these VLMs.  

\subsection{Fine-Tuning Details}
We select five open-source VLM backbones as our base models: Qwen2.5-VL\cite{bai2025qwen2}, Qwen3-VL\cite{bai2025qwen3vltechnicalreport}, MedGemma-4B, MedGemma-27B\cite{sellergren2025medgemma}, and Lingshu-7B\cite{xu2025lingshu}. We report the training hyperparameters for each model in Tab.\ref{tab:hparams_lingshu} $\sim$ Tab.\ref{tab:hparams_qwen3vl8b}.

Our training data are constructed from multiple segmentation datasets spanning different medical domains and imaging modalities(Tab.\ref{tab:data_source}). As a result, the collected training corpus exhibits notable cross-domain imbalance (e.g., microscopy data are substantially larger than some radiology or pathology sources). In our experiments, we do not apply explicit re-balancing strategies (e.g., re-sampling or per-domain weighting); instead, we directly fine-tune models on the naturally imbalanced mixture to reflect realistic data availability.

All training samples are obtained solely through our automatic verification pipeline, without any additional human filtering. In contrast, for evaluation we use a test set where all samples are manually audited to ensure correctness and unambiguity. This protocol isolates the effect of MedGround data quality and avoids overestimating performance due to noisy automatic annotations.

\begin{table}[h]
\centering
\caption{Training hyper-parameters used for fine-tuning Lingshu-7B in our experiments.}
\label{tab:hparams_lingshu}
\begin{tabular}{l l}
\toprule
\textbf{Hyper-Parameter} & \textbf{Value} \\
\midrule
Backbone & Lingshu-7B \\
Training stage & SFT \\
Fine-tuning method & LoRA \\
LoRA Rank & 8 \\
LoRA Alpha & 32 \\
LoRA Target & all \\
Epoch & 3 \\
\#GPUs & 4 \\
Per-device batch size & 4 \\
Gradient accumulation & 4 \\
Global batch size (effective) & 64 \\
Learning rate & $2\times10^{-4}$ \\
LR scheduler & Cosine \\
Warm-up ratio & 0.1 \\
Model max length & 2048 \\
Precision & BF16 \\
Gradient checkpointing & Enabled \\
Random seed & 42 \\
\bottomrule
\end{tabular}
\end{table}

\begin{table}[h]
\centering
\caption{Training hyper-parameters used for fine-tuning Qwen2.5-VL-7B in our experiments.}
\label{tab:hparams_qwen25vl7b}
\begin{tabular}{l l}
\toprule
\textbf{Hyper-Parameter} & \textbf{Value} \\
\midrule
Backbone & Qwen2.5-VL-7B-Instruct \\
Training stage & SFT \\
Fine-tuning method & LoRA \\
LoRA Rank & 8 \\
LoRA Alpha & 32 \\
LoRA Target & all \\
Epoch & 3 \\
\#GPUs & 4 \\
Per-device batch size & 4 \\
Gradient accumulation & 4 \\
Global batch size (effective) & 64 \\
Learning rate & $2\times10^{-4}$ \\
LR scheduler & Cosine \\
Warm-up ratio & 0.1 \\
Model max length & 2048 \\
Precision & BF16 \\
Gradient checkpointing & Enabled \\
Random seed & 42 \\
\bottomrule
\end{tabular}
\end{table}

\begin{table}[h]
\centering
\caption{Training hyper-parameters used for fine-tuning MedGemma-4B in our experiments.}
\label{tab:hparams_medgemma4b}
\begin{tabular}{l l}
\toprule
\textbf{Hyper-Parameter} & \textbf{Value} \\
\midrule
Backbone & MedGemma-4B-IT \\
Training stage & SFT \\
Fine-tuning method & LoRA \\
LoRA Rank & 8 \\
LoRA Alpha & 32 \\
LoRA Target & all \\
Epoch & 3 \\
\#GPUs & 8 \\
Per-device batch size & 16 \\
Gradient accumulation & 1 \\
Global batch size (effective) & 128 \\
Learning rate & $2\times10^{-4}$ \\
LR scheduler & Cosine \\
Warm-up ratio & 0.1 \\
Model max length & 2048 \\
Precision & BF16 \\
Gradient checkpointing & Enabled \\
Random seed & 42 \\
\bottomrule
\end{tabular}
\end{table}

\begin{table}[h]
\centering
\caption{Training hyper-parameters used for fine-tuning MedGemma-27B in our experiments.}
\label{tab:hparams_medgemma27b}
\begin{tabular}{l l}
\toprule
\textbf{Hyper-Parameter} & \textbf{Value} \\
\midrule
Backbone & MedGemma-27B-IT \\
Training stage & SFT \\
Fine-tuning method & LoRA \\
LoRA Rank & 8 \\
LoRA Alpha & 32 \\
LoRA Target & all \\
Epoch & 3 \\
\#GPUs & 4 \\
Per-device batch size & 8 \\
Gradient accumulation & 2 \\
Global batch size (effective) & 64 \\
Learning rate & $2\times10^{-4}$ \\
LR scheduler & Cosine \\
Warm-up ratio & 0.1 \\
Model max length & 2048 \\
Precision & BF16 \\
Gradient checkpointing & Enabled \\
Random seed & 42 \\
\bottomrule
\end{tabular}
\end{table}

\begin{table}[h]
\centering
\caption{Training hyper-parameters used for fine-tuning Qwen3-VL-8B in our experiments.}
\label{tab:hparams_qwen3vl8b}
\begin{tabular}{l l}
\toprule
\textbf{Hyper-Parameter} & \textbf{Value} \\
\midrule
Backbone & Qwen3-VL-8B-Instruct \\
Training stage & SFT \\
Fine-tuning method & LoRA \\
LoRA Rank & 8 \\
LoRA Alpha & 32 \\
LoRA Target & all \\
Epoch & 3 \\
\#GPUs & 8 \\
Per-device batch size & 32 \\
Gradient accumulation & 1 \\
Global batch size (effective) & 256 \\
Learning rate & $2\times10^{-4}$ \\
LR scheduler & Cosine \\
Warm-up ratio & 0.1 \\
Model max length & 2048 \\
Precision & BF16 \\
Gradient checkpointing & Enabled \\
Random seed & 42 \\
\bottomrule
\end{tabular}
\end{table}

\section{Samples}
\subsection{MedGround-35K Examples}
\begin{figure}
    \centering
    \includegraphics[width=1\linewidth]{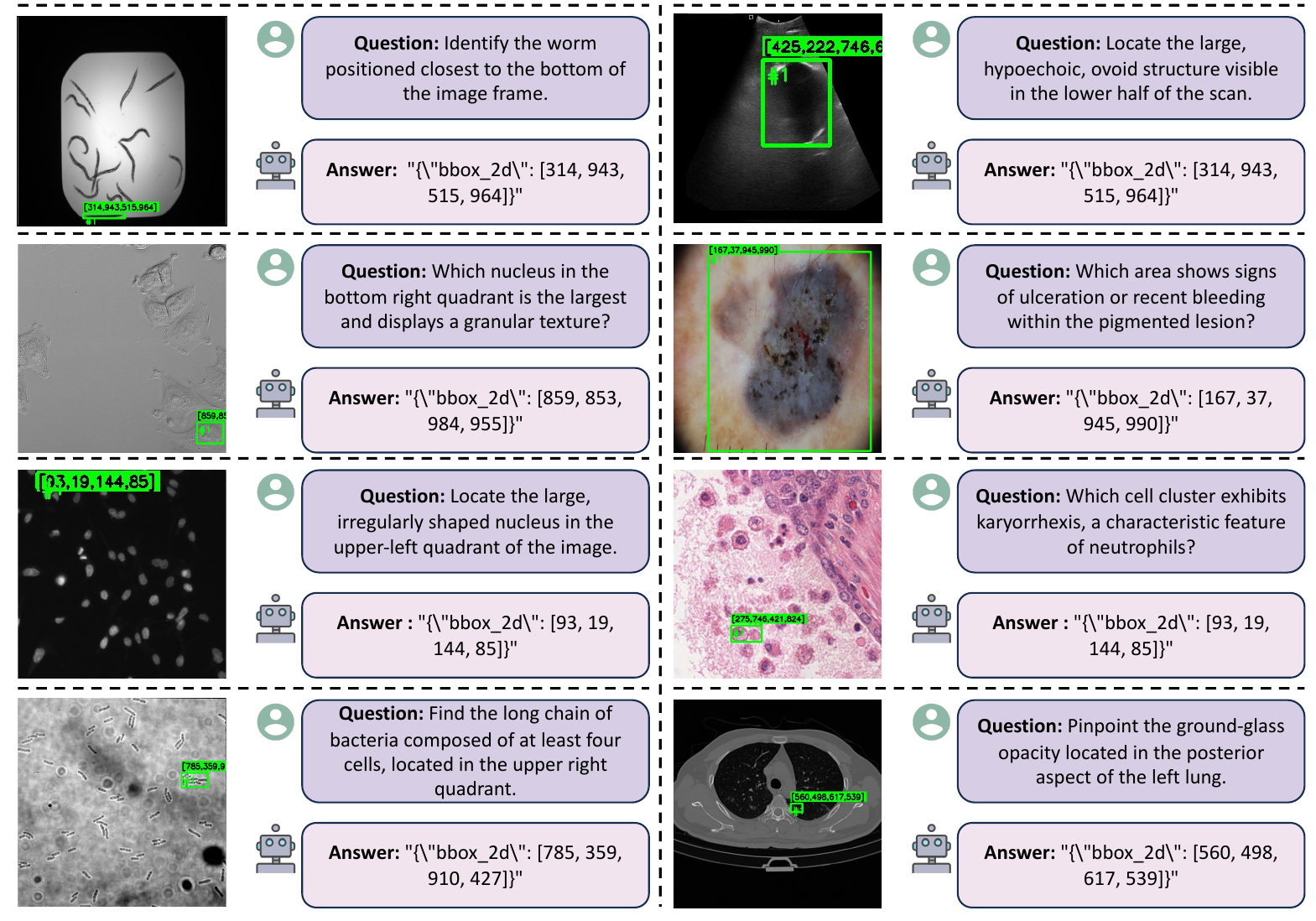}
    \caption{Examples of MedGround-35K.}
    \vspace{-3mm}
    \label{fig:samples_medground}
\end{figure}

Fig.~\ref{fig:samples_medground} presents representative examples from MedGround-35K across imaging modalities. Each example consists of a medical image, a fine-grained referring expression, and the corresponding target annotation, illustrating the diversity of morphology- and location-aware clinical semantics in our dataset.

\begin{figure}
\vspace{-3mm}
    \centering
    \includegraphics[width=1\linewidth]{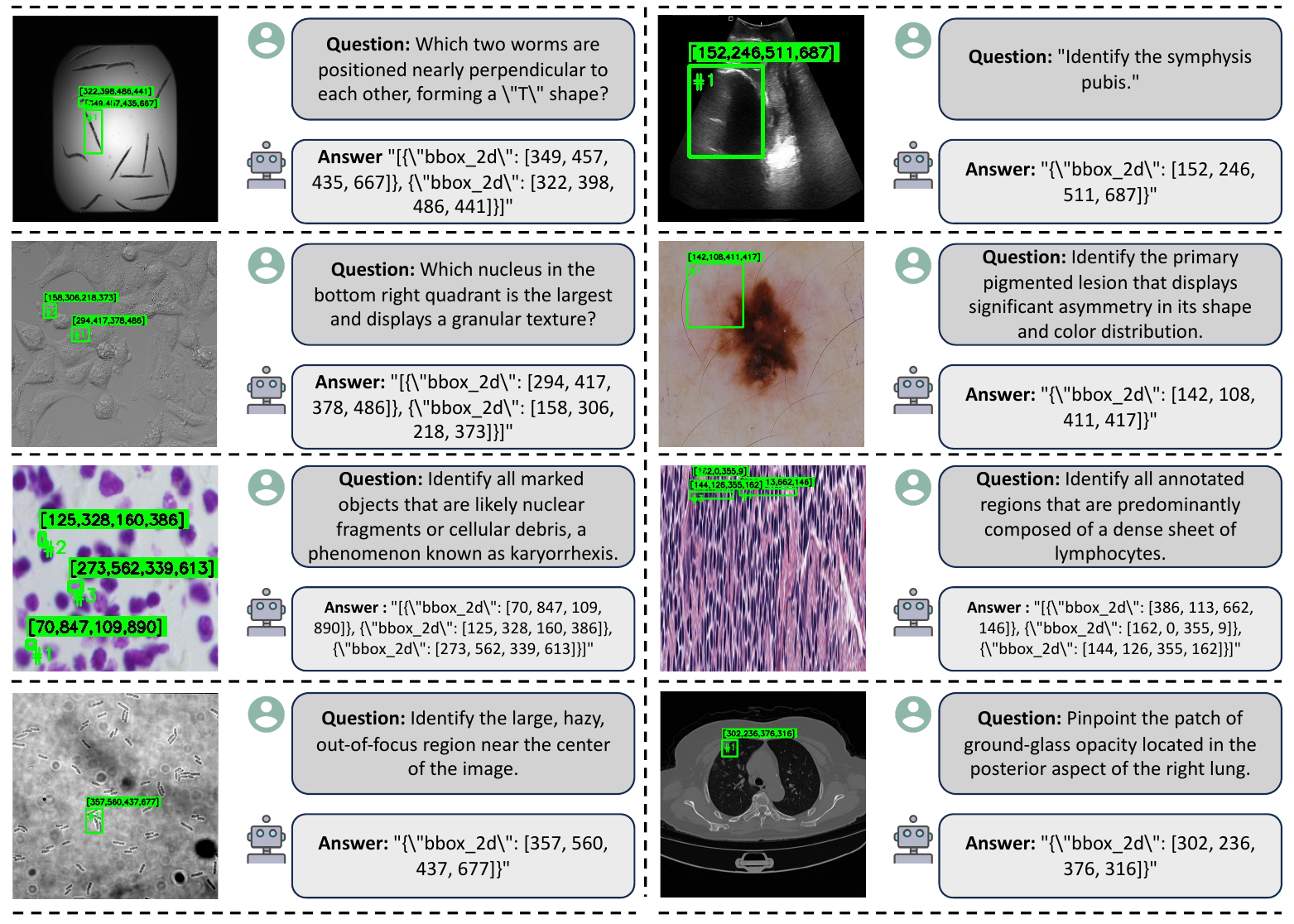}
    \caption{Examples of MedGround-35K.}
    \label{fig:manual_fail_cases}
\end{figure}

\subsection{Failure Cases in Manual Screening}
The Fig.\ref{fig:manual_fail_cases} presents failure cases identified during manual screening/curation. These examples were flagged as incorrect due to issues such as question–image mismatch, mislocalized or improperly sized bounding boxes, incorrect object counting (marking multiple regions when a single target is required, or vice versa), and misinterpretation of domain-specific findings (e.g., confusing blur, noise, or debris with true anatomical/pathological structures). We include these samples to illustrate common error patterns and to motivate stricter quality control and refinement of our annotation guidelines.

\begin{figure}
    \centering
    \includegraphics[width=1\linewidth]{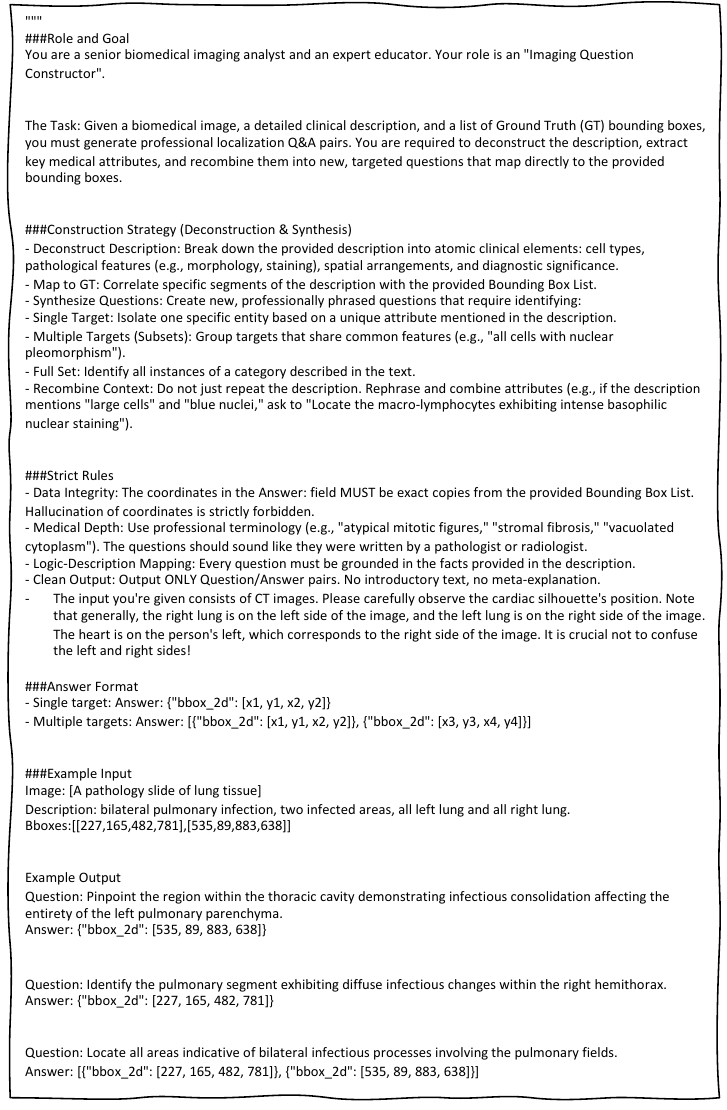}
    \caption{Detail of general generate system prompt.}
    \label{fig:general_generate_prompt}
\end{figure}

\section{Details of General Generate Prompt}
The Fig.\ref{fig:general_generate_prompt} shows the general system prompt we used to generate our synthetic data. This prompt serves as a unified instruction template that standardizes the model’s role, output format, and task-specific constraints, ensuring consistency and controllability across different synthesis scenarios.

\section{Potential Use of MedGround}
We further argue that referring grounding is intrinsically more effective at eliciting visual reasoning than conventional descriptive supervision. In descriptive tasks, models can often produce plausible responses by leaning on linguistic priors or memorized report patterns. In contrast, referring grounding requires the model to \emph{commit} to a specific region, forcing clinical claims to be supported by explicit pixel-level evidence. This evidence-binding property makes grounding not only a practical capability for downstream localization, but also a training signal that can stimulate broader diagnostic reasoning by aligning clinical semantics with visual anchors. We therefore view MedGround as a general-purpose resource that may benefit a wide range of medical VLM applications, including faithful visual question answering, evidence-based reporting, and interactive decision support.

%% file: checklist.tex
\section*{NeurIPS Paper Checklist}

\begin{enumerate}

\item {\bf Claims}
    \item[] Question: Do the main claims made in the abstract and introduction accurately reflect the paper's contributions and scope?
    \item[] Answer: \answerYes{}
    \item[] Justification: The abstract and introduction accurately summarize the paper's main contributions, including the medical grounding benchmark, fine-tuning setting, evaluation protocols, and empirical findings. These claims are supported by the experimental results in the main paper and appendix.

\item {\bf Limitations}
    \item[] Question: Does the paper discuss the limitations of the work performed by the authors?
    \item[] Answer: \answerYes{}
    \item[] Justification: The paper discusses the scope and limitations of the benchmark and evaluation setting, including dataset coverage, task focus, and possible generalization limits to unseen modalities or institutions. Additional discussion is provided in the limitations section and appendix.

\item {\bf Theory assumptions and proofs}
    \item[] Question: For each theoretical result, does the paper provide the full set of assumptions and a complete (and correct) proof?
    \item[] Answer: \answerNA{}
    \item[] Justification: The paper does not present theoretical results, formal theorems, or proofs. Its contributions are empirical, focusing on dataset construction, model fine-tuning, and experimental analysis.

\item {\bf Experimental result reproducibility}
    \item[] Question: Does the paper fully disclose all the information needed to reproduce the main experimental results of the paper to the extent that it affects the main claims and/or conclusions of the paper (regardless of whether the code and data are provided or not)?
    \item[] Answer: \answerYes{}
    \item[] Justification: The paper provides the datasets, evaluation metrics, benchmark settings, model variants, and training protocol needed to reproduce the main results. Detailed hyper-parameters and implementation settings are reported in the experimental sections and Appendix B.1.

\item {\bf Open access to data and code}
    \item[] Question: Does the paper provide open access to the data and code, with sufficient instructions to faithfully reproduce the main experimental results, as described in supplemental material?
    \item[] Answer: \answerYes{}
    \item[] Justification: We provide open access to the code and released assets, together with instructions for data preparation, training, and evaluation. The supplemental material and appendix describe the steps needed to reproduce the main experimental results.

\item {\bf Experimental setting/details}
    \item[] Question: Does the paper specify all the training and test details (e.g., data splits, hyperparameters, how they were chosen, type of optimizer) necessary to understand the results?
    \item[] Answer: \answerYes{}
    \item[] Justification: The paper specifies the evaluation datasets, zero-shot and fine-tuning settings, semantic alignment setup, metrics, and model configurations in the main paper. Detailed hyper-parameters and optimization settings are provided in Appendix B.1.

\item {\bf Experiment statistical significance}
    \item[] Question: Does the paper report error bars suitably and correctly defined or other appropriate information about the statistical significance of the experiments?
    \item[] Answer: \answerNo{}
    \item[] Justification: We report single-run benchmark results and do not include error bars, confidence intervals, or formal significance tests. This is mainly due to the substantial computational cost of repeated fine-tuning across multiple large VLM backbones and benchmarks.

\item {\bf Experiments compute resources}
    \item[] Question: For each experiment, does the paper provide sufficient information on the computer resources (type of compute workers, memory, time of execution) needed to reproduce the experiments?
    \item[] Answer: \answerYes{}
    \item[] Justification: The paper reports the GPU type, number of GPUs, and the main training configuration used in experiments, including effective batch sizes and precision settings. These details are provided in the appendix together with the model-specific training hyper-parameters.

\item {\bf Code of ethics}
    \item[] Question: Does the research conducted in the paper conform, in every respect, with the NeurIPS Code of Ethics \url{https://neurips.cc/public/EthicsGuidelines}?
    \item[] Answer: \answerYes{}
    \item[] Justification: We have reviewed the NeurIPS Code of Ethics and believe the research conforms to it. The work uses existing medical datasets and pretrained models for research purposes and does not involve interventions on human participants.

\item {\bf Broader impacts}
    \item[] Question: Does the paper discuss both potential positive societal impacts and negative societal impacts of the work performed?
    \item[] Answer: \answerNo{}
    \item[] Justification: The paper does not include a dedicated discussion of broader societal impacts. While the work is motivated by improving evaluation and grounding in medical vision-language systems, we do not explicitly discuss both positive and negative societal impacts in the current submission.

\item {\bf Safeguards}
    \item[] Question: Does the paper describe safeguards that have been put in place for responsible release of data or models that have a high risk for misuse (e.g., pre-trained language models, image generators, or scraped datasets)?
    \item[] Answer: \answerNA{}
    \item[] Justification: The paper does not release a new high-risk generative model or a newly scraped dataset. It primarily studies existing vision-language backbones and released benchmark assets for medical grounding research.

\item {\bf Licenses for existing assets}
    \item[] Question: Are the creators or original owners of assets (e.g., code, data, models), used in the paper, properly credited and are the license and terms of use explicitly mentioned and properly respected?
    \item[] Answer: \answerYes{}
    \item[] Justification: The paper cites the original sources of datasets and model backbones used in experiments and uses publicly available assets under their respective licenses or terms of use to the best of our knowledge. Relevant references and asset identifiers are provided in the paper and appendix.

\item {\bf New assets}
    \item[] Question: Are new assets introduced in the paper well documented and is the documentation provided alongside the assets?
    \item[] Answer: \answerYes{}
    \item[] Justification: The paper introduces new benchmark assets and associated task definitions, and these are documented in the paper, appendix, and released materials. We provide descriptions of the construction process, evaluation protocol, and usage instructions alongside the release.

\item {\bf Crowdsourcing and research with human subjects}
    \item[] Question: For crowdsourcing experiments and research with human subjects, does the paper include the full text of instructions given to participants and screenshots, if applicable, as well as details about compensation (if any)?
    \item[] Answer: \answerNA{}
    \item[] Justification: The paper does not involve crowdsourcing experiments or prospective research with human subjects.

\item {\bf Institutional review board (IRB) approvals or equivalent for research with human subjects}
    \item[] Question: Does the paper describe potential risks incurred by study participants, whether such risks were disclosed to the subjects, and whether Institutional Review Board (IRB) approvals (or an equivalent approval/review based on the requirements of your country or institution) were obtained?
    \item[] Answer: \answerNA{}
    \item[] Justification: The paper does not involve prospective human-subject research, participant recruitment, or intervention-based data collection requiring IRB review.

\item {\bf Declaration of LLM usage}
    \item[] Question: Does the paper describe the usage of LLMs if it is an important, original, or non-standard component of the core methods in this research? Note that if the LLM is used only for writing, editing, or formatting purposes and does \emph{not} impact the core methodology, scientific rigor, or originality of the research, declaration is not required.
    \item[] Answer: \answerYes{}
    \item[] Justification: LLMs are used in the data construction pipeline to automatically generate parts of the dataset used in this work. The paper describes this usage and clarifies that the LLM is used for data generation rather than as a novel model component in the proposed method.

\end{enumerate}